\documentclass[pdflatex,sn-mathphys-num]{sn-jnl}

\usepackage{graphicx}%
\usepackage{multirow}%
\usepackage{amsmath,amssymb,amsfonts}%
\usepackage{amsthm}%
\usepackage{mathrsfs}%
\usepackage[title]{appendix}%
\usepackage{xcolor}%
\usepackage{textcomp}%
\usepackage{manyfoot}%
\usepackage{booktabs}%
\usepackage{algorithm}%
\usepackage{algorithmicx}%
\usepackage{algpseudocode}%
\usepackage{listings}%

\usepackage[table]{xcolor} 
\usepackage{afterpage}
\usepackage{algcompatible}
\usepackage[title]{appendix}
\usepackage{caption}
\usepackage{enumitem}
\usepackage{etoolbox}
\usepackage[T1]{fontenc}
\usepackage{graphicx}
\usepackage{hyperref}
\usepackage[utf8]{inputenc}
\usepackage[switch]{lineno}
\usepackage{mathrsfs}
\usepackage{microtype}
\usepackage{minted}
\usepackage{multirow}
\usepackage{nicefrac}
\usepackage{pifont}
\usepackage{soul}
\usepackage{subcaption}
\usepackage{tcolorbox}
\usepackage{textcomp}
\usepackage{times}
\usepackage{multirow}
\usepackage{hyperref}

\newcommand{\cmark}{\ding{51}}%
\newcommand{\xmark}{\ding{55}}%

\theoremstyle{thmstyleone}%
\theoremstyle{thmstyletwo}%

\theoremstyle{thmstylethree}%

\begin{document}

\title[Article Title]{Automated Machine Learning for Unsupervised Tabular Tasks}

\author*[1]{\fnm{Prabhant} \sur{Singh}}\email{p.singh@tue.nl}
\author[1]{\fnm{Pieter} \sur{Gijsbers}}
\author[1]{\fnm{Elif Ceren} \sur{Gok Yildirim}}
\author[1]{\fnm{Murat Onur} \sur{Yildirim}}
\author[1]{\fnm{Joaquin} \sur{Vanschoren}}



\affil[1]{\orgdiv{AMOR/e Lab}, \orgname{Eindhoven University of Technology}, \orgaddress{ \city{Eindhoven}, \postcode{5600 MB}, \country{Netherlands}}}


\maketitle

\begin{abstract}

In this work, we present LOTUS (Learning to Learn with Optimal Transport for Unsupervised Scenarios), a simple yet effective method to perform model selection for multiple unsupervised machine learning(ML) tasks such as outlier detection and clustering. Our intuition behind this work is that a machine learning pipeline will perform well in a new dataset if it previously worked well on datasets with a \textit{similar} underlying data distribution . We use Optimal Transport distances to find this similarity between unlabeled tabular datasets and recommend machine learning pipelines with one unified single method on two downstream unsupervised tasks: outlier detection and clustering. We present the effectiveness of our approach with experiments against strong baselines and show that LOTUS is a very promising first step toward model selection for multiple unsupervised ML tasks.\footnote{Accepted at Machine Learning Journal, ACML 2026 Special Track }

\end{abstract}


\section{Introduction}
Automated Machine Learning (AutoML)~\cite{Hutter2019AutomatedML} aims to automate the design and optimization of machine learning pipelines in a data-driven way, using a variety of optimization techniques to find the best pipeline in a vast search space of possible pipelines consisting of many data preparation steps and modeling techniques. AutoML has shown promising results in supervised settings like classification~\cite{Hutter2019AutomatedML}, regression, and forecasting~\cite{Shchur2023AutoGluonTimeSeriesAF}. Several AutoML techniques rely on optimization like Bayesian optimization~\cite{NIPS2015_11d0e628} and evolutionary search~\cite{gama} to achieve these results. Many AutoML~\cite{Hutter2019AutomatedML} tools leverage meta-learning schemes~\cite{Vanschoren2018MetaLearningAS} to find good configurations to warm-start optimization. For instance, AutoSklearn-2.0~\cite{feurer-arxiv20a} learns pipeline portfolios, FLAML~\cite{Wang2021FLAMLAF} uses meta-learned defaults. However, for unsupervised tasks i.e. the tasks that lack access to ground truth labels, the effectiveness of AutoML tools is very limited because of the lack of evaluation metrics during search and optimization.

\subsection{AutoML for Unsupervised tasks}

Recent works in automated model selection for outlier detection~\cite{metaod, anonymous2024hpod, TautoOD} use meta-learning to recommend outlier detection algorithms that perform well on similar tasks, where task similarity is estimated using a subset of meta-features that do not require labels, especially landmarks and model-based meta-features. In MetaOD~\cite{metaod} a collaborative filtering (CF) technique~\cite{stern2010collaborative} is used to recommend algorithms for a given task. PyODDS~\cite{Li2020PyODDSAE} is a related method but it requires ground truth data to select specific outlier detection pipelines.  One can argue that the use of internal metrics such as Excess-Mass~\cite{emmv}, Mass-Volume~\cite{emmv}, and IREOS~\cite{IREOS} can be used for model selection on outlier detection tasks instead.  However, it has been shown that these internal metrics for outlier detection algorithms are computationally extremely expensive and do not scale well to large datasets~\cite{internalstrategysurvey} and show little or no correlation with external metrics. AutoML for clustering follows similar works which utilize a combination of internal metrics, meta-learned features, and optimization~\cite{ML2DAC, autocluster, clusteringapp, liu2021autocluster, poulakis2020autoclust, tschechlov2021automl4clust}, in clustering evaluation metrics usually referred to as Cluster Validity Indices(CVI). We usually have task-specific tools for unsupervised tasks like clustering and outlier detection. These tools use task-specific model-based meta-features to give optimal pipelines. We compare some of these tools with LOTUS in Table \ref{tab:overview}.

\begin{table}[h]
\small
\setlength{\tabcolsep}{2.5pt}
\begin{tabular}{cccc}
\hline
                                                & \textbf{Meta-Learning}      & \textbf{Outlier detection} & \textbf{Clustering} \\ \hline
MetaOD~\cite{metaod}                              & Collaborative filtering                                   & \xmark                          & \cmark \\
AutoML4Clust~\cite{tschechlov2021automl4clust} & \xmark                                       & \xmark                         & \cmark \\
AutoClust~\cite{poulakis2020autoclust}            &  CVI                                       & \xmark                         & \cmark \\
AutoCluster~\cite{liu2021autocluster}             &  CME, CVI                                  & \cmark                         & \xmark \\
LOTUS(Ours)                                             &   OT                                       & \cmark                         & \cmark \\ \hline
\end{tabular}
\caption{Overview of prior work on automated clustering and automated outlier detection, indicating which components of model selection and hyperparameter optimization are addressed by each method.}
\label{tab:overview}
\end{table}

\subsection{Our Method}
In this work we generalize and extend our previously published work on automated machine learning for outlier detection~\cite{ijcai2023p843}. We propose LOTUS, a two-phase meta-learning method for model selection for multiple unsupervised machine learning tasks that leverage optimal transport distances to recommend which unsupervised algorithms, preprocessing techniques, and hyperparameters to use based on how well they performed on prior tasks with similar data distributions. LOTUS first transforms the datasets and then finds the most similar dataset from the meta-dataset and recommends the optimal algorithm. In this work, we aim to contribute to a scenario where the user requires an unsupervised algorithm for a given task where one does not have availability to the labels for a given task but has access to labels for previously evaluated tasks. This work evaluates our approach for model selection on outlier detection and clustering tasks. The key contributions of this work are:

\begin{enumerate}
    \item \textbf{LOTUS}: A meta-learning technique based on finding similar tasks using Optimal Transport for unsupervised scenarios.
    \item Experimental evaluation of LOTUS with strong baselines with extensive experiments on unsupervised tasks (Clustering and Outlier Detection), demonstrating that LOTUS yields significantly better results.
    \item Additionally, with LOTUS we provide two open source AutoML systems that can perform supervised model selection for outlier detection and clustering. Our code is available on \url{https://github.com/prabhant/LOTUS-CL-OD}
\end{enumerate}

\section{Preliminary: Optimal Transport} \label{prelim:ot}
Optimal transport (OT) or transportation theory, also known as Kantorovich–Rubinstein duality, is a problem that deals with the transportation of masses from source to target~\cite{Villani2008OptimalTO}. This problem is also called the Monge–Kantorovich transportation problem~\cite{Villani2008OptimalTO}. In recent years, OT has gained significant attention from the machine learning community, as it provides a powerful framework for designing algorithms that can learn to match two probability distributions. In this section, we give an introduction to OT and distance measures related to our work. 

In OT, the objective is to minimize the cost of transportation between two probability distributions. For a cost function between pairs of points, we calculate the cost matrix $C$ with dimensionality $n \times m $. The OT problem minimizes the loss function $L_c(P):=\langle C, P\rangle$ with respect to a coupling matrix $P$. A practical and computationally more efficient approach is based on regularization and minimizes $L_c^\epsilon(P):=\langle C,P\rangle + \epsilon \cdot r(P)$ where $r$ is the negative entropy, computed by the Sinkhorn algorithm~\cite{Cuturi2013SinkhornDL}, and $\epsilon$ is a hyperparameter controlling the amount of regularization. A discrete OT problem can be defined with two finite point clouds, 
$\{x^{(i)}\}^{n}_{i=1}$ ,$\{y^{(j)}\}^{m}_{j=1}, x^{(i)},y^{(j)}\in \mathbb{R}^d $, which can be
described as two empirical distributions: $\mu:=\sum^n_{i=1}a_i\delta_{x^{(i)}}, \nu:=\sum^m_{j=1}b_j\delta_{y^{(j)}}$. 
Here, $a$ and $b$ are probability vectors of size $n$ and $m$, respectively, and the $\delta$ is the Dirac delta.

\section{Method: Learning to learn with Optimal Transport for Unsupervised Scenarios} \label{methodssection}

We introduce LOTUS: \textbf{L}earning to learn with \textbf{O}ptimal \textbf{T}ransport for \textbf{U}nsupervised \textbf{S}cenarios. Unsupervised tasks, by definition, lack ground-truth labels on new data, rendering direct model optimization infeasible. We overcome this by meta-learning from prior experiences for which we do have a ground truth and then we transfer that knowledge to new, unlabeled tasks. Hence, this strategy is realized through a two-phase system. First we learn from data collected in previous tasks and then recommends a pipeline for a new unseen task. We call these two phases \textit{meta-training} and \textit{model selection} respectively. The meta-training phase is intended to fill the population of our meta-dataset with optimal pipelines searched on historically sampled datasets. In the model selection phase, LOTUS finds the most similar dataset to the current dataset and transfers the optimal pipelines with hyperparameter configuration to that pipeline.

\subsection{Meta-training phase}\label{subsec: metatrain}
In this section first we formally introduce our problem of meta-training. In the first phase, LOTUS meta-learns how well different unsupervised algorithms work on prior \textit{labeled} datasets. These can be datasets where the correct labels are known or proxy tasks. More formally, we require a collection of $n$ prior labeled datasets $\mathcal{D}_{meta} = \{D_1, ..., D_n\}$ with train and test splits such that $D_i = (X^{train}_i, y_i^{train}),(X_i^{test}, y_i^{test})$. The result of meta-training is collection of $m$ optimized pipelines $A^{*}_i$ with associated hyperparameters $\lambda^{*}_i$ for every dataset in $\mathcal{D}_{meta}$; $\mathcal{A} = \{A^*_{\lambda^*_1}, ..., A^*_{\lambda^*_m}\}$. 

\textbf{Problem Formulation:} The meta-training phase of LOTUS acts like a typical  Combined Algorithm Selection and Hyperparameter optimization (CASH) problem, stated in equation \ref{eq:cash}, where $A_{\lambda^*}^*$ is the combination of the optimal algorithm from search space $A$ with associated hyperparameter space $\Lambda_A$ evaluated over $k$ cross-validation folds of dataset $D=\{X,y\}$ with training and validation splits. $L$ is our evaluation measure.
\begin{equation}
\begin{split}
A^*_{\lambda^*} =
\operatorname*{argmin}_{%
       \substack{%
         \forall A^j \in \boldsymbol{A} \\
         \forall \lambda \in \boldsymbol{\Lambda_A}
       }
     }
\frac{1}{k} 
\sum_{f=1}^{k} L \left( A^j_\lambda, \big\{\boldsymbol{X}^{train}_f, \boldsymbol{y}^{train}_f\big\}, \big\{\boldsymbol{X}^{val}_f, \boldsymbol{y}^{val}_f\big\} \right)
\end{split}
\label{eq:cash}
\end{equation}
\par
The CASH problem from Equation \ref{eq:cash} relies on the validation split to optimize for the optimal configuration. However, in unsupervised settings, such validation splits are not relevant. We run estimators on all unlabeled data and use the ground truth labels only to evaluate them. Our modified CASH formulation to select the optimal unsupervised algorithm \textbf{with access to labels} is as follows:

\begin{equation}
\begin{split}
A^*_{\lambda^*} =
\operatorname*{argmin}_{%
       \substack{%
         \forall A^j \in \boldsymbol{A} \\
         \forall \lambda \in \boldsymbol{\Lambda_A}
       }
     }
 L \left( A^j_\lambda, \big\{\boldsymbol{X}\} \big\{\boldsymbol{y}\big\} \right)
\end{split}
\label{eq: cashunsupervised}
\end{equation}
\par
Note that this CASH formulation is only applied for populating our knowledge base $\mathcal{A}$ and we do not expect labels for new datasets while model selection. 

To show how LOTUS meta-training works we present Algorithm \ref{algo: LOTUS train}. In Algorithm \ref{algo: LOTUS train} first a dataset is selected from the meta-dataset and then we use the optimization strategy to find the optimal pipelines from our predefined search space based on our optimization metric $L$ for the given dataset. We return the optimal pipeline for the given dataset and add $A^*_{\lambda^*i}$ to our meta-dataset of optimal algorithms $\mathcal{A}$. We describe the task-specific changes for clustering and outlier detection in Section \ref{taskspecificextension} where we elaborate on the meta-training setup for the clustering and outlier detection tasks.

\begin{algorithm}[tb]
    \caption{Algorithm for Meta-training; For each dataset $D_i$ in $\mathcal{D}_{meta}$ an optimization strategy is employed to find the optimal pipeline $A^*_{\lambda^*i}$ }\label{algo: LOTUS train}
    \begin{flushleft}
        \textbf{Inputs:} $\mathcal{D}_{meta}, L, \boldsymbol{A},\boldsymbol{\Lambda}_{\boldsymbol{A}}$ \Comment{Meta-datasets, evaluation measure, models and hyperparameters}
    \end{flushleft}
    \begin{algorithmic}[1]
            \WHILE{$D_{i} \in \mathcal{D}_{meta}$}
            \STATE $A^*_{\lambda^*i} \gets \operatorname*{argmin}_{%
       \substack{%
         \forall A^j \in \boldsymbol{A} \\
         \forall \lambda \in \boldsymbol{\Lambda_A}
       }
     }
 L \left( A^j_\lambda, \big\{\boldsymbol{X}\} \big\{\boldsymbol{y}\big\} \right)
     $
         \STATE $\mathcal{A} \gets A^*_{\lambda^*i}$

            \ENDWHILE
    \end{algorithmic}
\end{algorithm}

\subsection{Model Selection Phase}\label{modelselectionphase}

In the second phase, given a new input dataset $D_{new} = ( X_{new} )$ without any labels, we aim to select a pipeline $A^*_{\lambda^*} \in \mathcal{A}$ to employ on $X_{new}$, where $A^*_{\lambda^*} $ is a tuned pipeline for a dataset similar to $X_{new}$. Our premise is that, if a prior dataset exists that is very similar to the new dataset, then its optimal pipelines will likely work well on the new dataset. In the following section we motivate why this premise is valid given appropriate choices of distance and preprocessing methods.

\subsubsection{Finding a distance for Tabular datasets}

\begin{figure*}[th]
\begin{center}
\includegraphics[height=6.3cm]{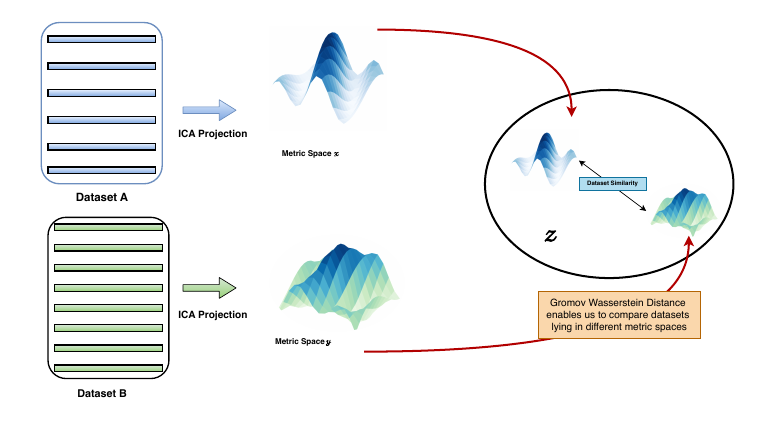}
\end{center}
\caption{Figure demonstrating our approach to measure dataset similarity between two datasets from different domains.}\label{fig: gwlr}
\end{figure*}


In this section we discuss the the use of Optimal Transport (as described in Section \ref{prelim:ot}) to compute distances between unlabeled tabular datasets of different sizes. To compute OT distances we treat the datasets as measure metric spaces (mm-space). 
While the Wasserstein distance can be used in this context, it has limitations in tabular data from different domains. Specifically, if two datasets  differ substantially in their feature types such as one originating from a biology domain and the other from physics their features may not be directly comparable. This results in Wasserstein distance producing misleading results, since it assumes a shared metric space between features, which does not hold in this case. \textit{We need to find a distance which allows us to measure similarity between datasets with different sizes and different domains.}

In this work we propose using Gromov-Wasserstein distance~\cite{Mmoli2011GromovWassersteinDA} to address the issues mentioned above. Gromov-Wasserstein distance offers a powerful alternative as it does not require features to reside in a shared metric space or possess direct comparability. Instead, it operates by comparing the internal metric structures of the datasets (i.e., how points relate to each other within each dataset, captured by their respective intra-dataset distance matrices), making it inherently suitable for comparing datasets with disparate feature sets. Our work builds on the mathematical properties of Gromov–Wasserstein(GW) distance and its relationship to unsupervised learning algorithms. We draw inspiration from extrinsic and intrinsic similarity between datasets shown by \citet{Mmoli2011GromovWassersteinDA} and the duality between clustering and sketching in metric measure spaces established by~\citet{mémoli2018sketchingclusteringmetricmeasure}. We visually describe our approach in Figure \ref{fig: gwlr}.

The goal of using this distance is not only to quantify dataset similarity but also to identify suitable unsupervised algorithms such as clustering or outlier detection for a given dataset. These algorithms often depend on the internal structure of the data, which Gromov-Wasserstein distance is designed to capture how points relate to each other and how mass is distributed. This makes it a more appropriate choice for tasks involving automated selection of algorithms in unsupervised tabular learning. For example performance of clustering and OD algorithms depend on similar factors.

The Gromov-Wasserstein distance compares the relational structure of distributions rather than relying on a strict one-to-one correspondence of features. This is especially suited for unsupervised machine learning scenarios, where structural properties play a big role in the performance of algorithm for the given task(clustering or OD).  This problem can be written as a function of $(a,A), (b,B)$ between our distributions $A$ and $B$~\cite{Villani2008OptimalTO,gwlr}:
\begin{equation}
        \text{GW}((a,A),(b,B)) =
        \min_{P\in \Pi_{a,b}} \mathcal{Q}_{A,B}(P)
\end{equation}
where $\Pi_{a,b}:=\{ P \in \mathbb{R}^{n \times m}_+| P\mathbf{1}_m = a, P^{T}\mathbf{1}_n=b\}$ is the set of all possible mappings of points from $A$ to $B$ and the \textit{energy} $\mathcal{Q}_{A,B}$ is a quadratic function of $P$  which can be described as 
\begin{equation}
    \mathcal{Q}_{A,B}(P):= \sum_{i,j,i^{'},j^{'}}(A_{i,i'}-B_{j,j'})^2P_{i,j}P_{i',j'}
    \label{eq: gwlr-np}
\end{equation} 

This distance comes with an overhead as computing Gromov-Wasserstein distance is NP-hard.
\subsubsection{Computational Consideration}

The NP-hardness of the GW distance (Equation \ref{eq: gwlr-np}) requires approximations for practical use. A common approach is entropic regularization of Gromov Wasserstein distance(Equation \ref{eq: entropicgwlr}) proposed by~\citet{pmlr-v48-peyre16} which adds an entropy term to the objective, thereby smoothing the problem and often facilitating faster computation.
\begin{equation}\label{eq: entropicgwlr}
    \text{GW}_\varepsilon((a,A),(b,B)) = \min_{P\in \Pi_{a,b}} \mathcal{Q}_{A,B}(P) - \varepsilon \cdot H(P)    
\end{equation}
where $GW_\epsilon$ is the Entropic Gromov Wasserstein cost between our distributions $A$ and $B$,
$H(P)$ is the Shannon entropy, and $\varepsilon$ a regularization constant. We now have a distance defined to find the similarity between two unlabeled tabular datasets(which is the case for clustering and outlier detection). 

For further scalability, particularly to achieve the linear time complexity desirable for large meta-datasets,  we use the Low-Rank approximation of Gromov Wasserstein (GW-LR) approximation~\cite{pmlr-v139-scetbon21a, Scetbon2022LowrankOT,gwlr}, which reduces the computational cost from cubic to linear time. ~\citet{gwlr} consider the Gromov Wasserstein problem with low-rank couplings, linked by a common marginal $g$. Therefore, the set of possible transport plans is restricted to those adopting the factorization of the form $P_r = Q_{diag}(1/g)R^T$, where $Q$ and $R$ are thin matrices with the dimensionality of $n\times r$ and $r\times m$, respectively, and $g$ is an $r$-dimensional probability vector. 
The GW-LR distance is then described as:
\begin{equation}\label{eq 2-2}
\vspace{5pt}
\begin{aligned}
    \text{GW-LR}^{(r)}  ((a,A), (b,B)) :=  \min_{(Q,R,g)\in \mathcal{C}_{a,b,r}}\mathcal{Q}_{A,B}(Q_{diag}(1/g)R^T)
\end{aligned}
\end{equation}

By using GWLR we can compute the similarity between two tabular datasets in with linear time complexity, but these tabular datasets do contain a degree of noise, for example having different range of parameters in every column and continuous values in one and categorical in another. This noise makes it hard for us to compare two datasets; in the next section, we talk about how to apply preprocessing to enable us to do that.

\subsubsection{Preprocessing}\label{subsec preprocessing}

In this work we aim to solve this problem for real-world datasets, these datasets comes with additional noise in them, for example they can contain non numerical values, categorical values, different ranges of values depending on one dataset to another. To make sure our similarity computation works on these real-world datasets we can use a preprocessing method to eliminate this noise from our datasets and make our dataset suitable for Gromov-Wasserstein distance. 

To ensure that the Gromov-Wasserstein distance captures fundamental structural similarities rather than superficial differences arising from feature scaling, inherent noise, or feature redundancy, we preprocess each dataset using Fast Independent Component Analysis (FastICA). FastICA is a widely used blind source separation algorithm that projects data onto a set of statistically independent components by maximizing non-Gaussianity~\cite{Hyvrinen2000IndependentCA}. FastICA transforms each dataset into a latent space that captures independent modes of variation intrinsic to the data, independent of the original feature semantics. By doing so, it aims to reveal a more intrinsic geometry of the data, less affected by the original feature representation. As Gromov-Wasserstein distance compares datasets based on their internal relational structures rather than on a shared feature space, applying FastICA as a preprocessing step produces a more meaningful and stable metric space for GW-based comparisons. This is particularly beneficial in unsupervised learning settings, where we aim to identify structurally similar datasets or recommend suitable algorithms without relying on labeled data.

For our preprocessing function, we have a dataset $D \in \mathbb{R}^{n \times m} $  where $n$ is the number of samples and $m$ is the number of features. We apply FastICA~\cite{Hyvrinen2000IndependentCA}  as our preprocessing algorithm.

Let \( \mathcal{F} \) denote the FastICA transformation:
\[
\mathcal{F}: \mathbb{R}^{n \times d} \rightarrow \mathbb{R}^{n \times k}
\]
which projects a dataset \( D \in \mathbb{R}^{n \times d} \) into a \(k\)-dimensional representation \( \mathbf{Z} \in \mathbb{R}^{n \times k} \) of statistically independent components.
For two datasets \( D_a \) and \( D_b \), we compute:
\[
\mathbf{Z}_a = \mathcal{F}(D_a), \quad \mathbf{Z}_b = \mathcal{F}(D_b)
\]

We then define the distance between datasets using the Gromov-Wasserstein distance combined with a low-rank approximation (GW-LR) as:
\begin{equation}
    \mathcal{O} = GW\text{-}LR^{(r)}\left( \mathcal{F}(D_a), \mathcal{F}(D_b) \right)
    \label{eq:LOTUSGWLR}
\end{equation}



The most similar prior dataset $D_{s} \in \mathcal{D}_{meta}$ is the dataset with the smallest distance to the new dataset $D_{new}$.
\begin{equation} 
   D_{s} = \underset{i}{\mathrm{argmin}}  \mathcal{O}_i
   \label{eq: similar dataset}
\end{equation}
LOTUS then assigns the optimal configuration from $\mathcal{A}$: $A^*_{\lambda^{*}_{new}} = A^*_{\lambda^{*}_s}$ where $A^*_{\lambda^{*}_s}$ is predicted as the optimal configuration for $D_{new}$. We describe this model selection phase in Algorithm \ref{algo: LOTUS}, where we get a new dataset $D_{new}$ as input to LOTUS. We iterate through every dataset in $\mathcal{D}_{meta}$. Once we find the dataset with the least distance from our meta-dataset we transfer the corresponding optimal pipeline to the new dataset.

\begin{algorithm}[tb]
    \caption{Algorithm for LOTUS }\label{algo: LOTUS}
    \begin{flushleft} 
    \textbf{Inputs:} $D_{new}, \mathcal{D}_{meta}, \mathcal{A}$
    \end{flushleft}
\begin{algorithmic}[1]
    \WHILE{$D_{i} \in \mathcal{D}_{meta}$}
    \STATE $\mathcal{O}_i \gets GWLR(\mathcal{F}(D_{new}),\mathcal{F}(D_i))$\COMMENT{Distance calculation}
    \ENDWHILE
    \STATE $s \gets {\mathrm{argmin}}\{\mathcal{O}_1,...,\mathcal{O}_n\}$\COMMENT{Retrieval of most similar dataset}
    \STATE $A^*_{\lambda^{*}_{new}} \gets A^*_{\lambda^{*}_s}$\COMMENT{$A^*_{\lambda^{*}_s}$ is the pipeline associated with the most similar dataset in $\mathcal{D}_{meta}$}
    \end{algorithmic}
\end{algorithm}

\subsection{Task-specific Implementations} \label{taskspecificextension}

\begin{figure*}[h]
  \centering
  \includegraphics[width=0.8\textwidth]{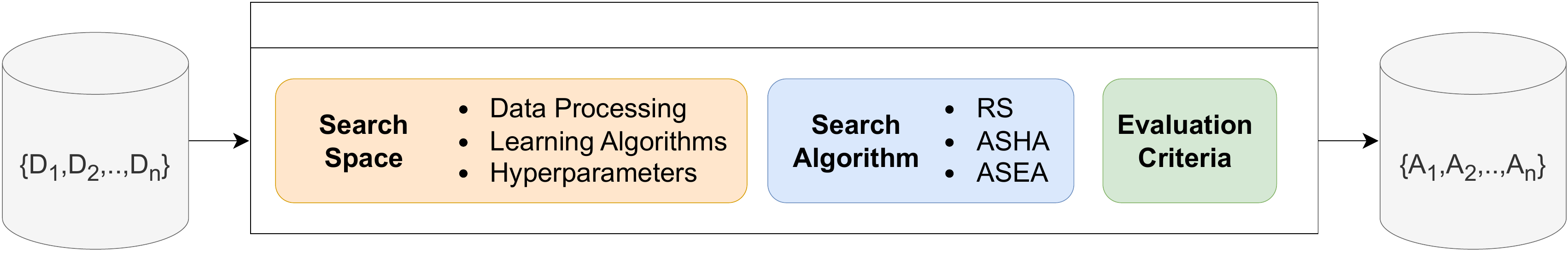}
  \caption{Figure demonstrating our extensions that populates $\mathcal{A} = \{A^*_{\lambda^*_1}, ..., A^*_{\lambda^*_n}\}$}    
  \label{fig: GAMAFig}

\end{figure*}

The Meta-training phase as formulated in Section \ref{subsec: metatrain} requires robust mechanism to discover optimal unsupervised pipelines $A_{\lambda^*}$ on historical datasets within $\mathcal{D}_{meta}$ and build $\mathcal{A}$ To implement this we developed task-specific AutoML systems LOTUS-Outlier for Outlier detection and LOTUS-Clust for clustering. Both systems are build upon the GAMA AutoML framework~\cite{Gijsbers2021} and are responsible for search the pipeline space and populating $\mathcal{A}$. We use a search strategy to iterate through search space and return an optimal pipeline and we then populate our algorithm store with these optimal pipelines as discussed in meta-training phase in Algorithm \ref{algo: LOTUS train}. In Figure \ref{fig: GAMAFig} the meta-training in our setup follows the following steps:

\begin{enumerate}
    \item Individual datasets $D_i$ from $\mathcal{D}$ are input to the our AutoML systems.
    \item We define a search space and select a search strategy that finds an optimal algorithm.
    \item We provide the evaluation criteria/metric to evaluate our pipelines.
\end{enumerate}
We call these extensions simply LOTUS-Outlier and LOTUS-Clust. We allow users to select either random search(RS), evolutionary algorithm(ASEA), and ASHA~\cite{asha} for searching the optimal pipeline and select from various metrics like F1, AMI, CH, ARI, etc.

\subsubsection{Clustering}
We developed a clustering based extension for our problem named LOTUS-Clust. If historical dataset includes labels then LOTUS-Clust can find optimal performing pipelines such by using an external CVI like Adjusted Mutual Information~\cite{AMI}. However, if a dataset does not have true labels, LOTUS-Clust can still identify pipelines that perform well by optimizing based on internal cluster validity indicators (CVI), such as the Calinski-Harabasz index~\cite{calinski1974dendrite}.  This latter scenario is formalized in Equation \ref{eq: internalmetricopt}.
\begin{equation}
\begin{split}
A^*_{\lambda^*} =
\operatorname*{argmin}_{%
       \substack{%
         \forall A \in \boldsymbol{A} \\
         \forall \lambda \in \boldsymbol{\Lambda_A}
       }
     }
 CVI \left( A_\lambda, \big\{\boldsymbol{X}\} \right)
\end{split}
\label{eq: internalmetricopt}
\end{equation}
\begin{table}[h]
\begin{center}
\begin{minipage}{\textwidth}
\begin{tabular}{@{}lll@{}}
\toprule
Algorithm($A_i$) & Hyperparameter  & Search Space($\Lambda_i$)\\
\midrule
k-Means    & n\_clusters   & [2-21]   \\
           & n\_init\   & ['auto']      \\
           & max-iter   & [300-500]      \\
           & algorithm & ['lloyd', 'elkan'] \\
MiniBatchKmeans  & n\_clusters   & [2, 21]   \\
                & max-iter   & [100-500]      \\
                & min-bin-freq & [1,2,3,4,5]\\
Mean Shift & bin\_seeding & [True, False] \\
           & min\_bin\_freq & [1-5]) \\
           & max\_iter & [2-300]      \\
AgglomerativeClustering & n\_clusters & n [2-21]   \\
           & affinity & [‘euclidean’, ‘manhattan’, ‘cosine’, ‘l1’, ‘l2’] \\
           & linkage & [‘ward’, ‘complete’, ‘average’, ‘single’] \\
DBSCAN & eps & [0.1-0.5]\\
       & min\_samples & [3,4,5,6,7,8] \\
       & p & [1, 2] \\
OPTICS & min\_samples & [3,4,5,6,7,8\\
       & p & [1, 2] \\
       & xi & [0.05-5] \\
BIRCH & n\_clusters   &  [2-21]   \\
      & threshold     & [0.2-0.8] \\
      & branching\_factor & [25, 50, 75] \\\hline
        
\end{tabular}
\caption{ Search spaces $\Lambda_i$ of hyperparameters for each clustering algorithm $A_i$ used in LOTUS-Clust. Search space includes centroid-based methods; density-based approaches like DBSCAN and OPTICS; hierarchical clustering via AgglomerativeClustering; and model-based methods like Mean Shift and BIRCH.}
\label{tab: searchspace}

\end{minipage}
\end{center}
\end{table}
\par

 Our search space uses the most popular clustering algorithms from Scikit-Learn~\cite{scikit-learn} as described in Table \ref{tab: searchspace}. The search space we use for LOTUS-Clust is inspired by previous works~\cite{poulakis2020autoclust, autocluster} which used a similar search space.

\subsubsection{Outlier Detection}
We develop LOTUS-Outlier for outlier detection in supervised settings. One can look at LOTUS-Outlier as a tool for algorithm selection and hyperparameter optimization in outlier detection settings where one has labels available for a part of the data. The optimization in LOTUS-Outlier is based on similar logic as Equation \ref{eq: cashunsupervised}. 
For outlier detection, we use the search space described in Table \ref{tab:gamaodsearchspace}, we use a similar search space used by MetaOD \footnote{We implement the same search space as MetaOD GitHub repository for a fair comparison. \url{https://github.com/yzhao062/MetaOD/blob/master/metaod/models/base\_detectors.py}, MetaOD also uses all the existing datasets from ADbench.}
\begin{table}[]
    \centering
        \begin{tabular}{lll}
        \toprule
        Algorithm( $A_i$) & Hyperparameter & Search Space($\Lambda_i$) \\
        \midrule
        LODA & n\_bins & [5, 10, 15, 20, 25, 30] \\
         & n\_random\_cuts & [10 - 200] \\
        ABOD & n\_neighbors & [3, 5, 10, 15, 20, 25, 50, 60, 75] \\
        IForest & n\_estimators & [10 - 200] \\
         & max\_features & [0.1 - 0.9] \\
        KNN & n\_neighbors & [ 1- 100] \\ 
         & method & ['largest', 'mean', 'median'] \\
        LOF & n\_neighbors & [1- 100] \\
         & metric & ['manhattan', 'euclidean', 'minkowski'] \\
        HBOS & n\_bins & [5 - 100] \\
         & alpha & [0.1 - 0.5] \\
        OCSVM & nu & [0.1 - 0.9] \\
         & kernel & ['linear', 'poly', 'rbf', 'sigmoid'] \\
        \bottomrule
        \end{tabular}
    \caption{Domain of Hyperparameters $\Lambda_i$ for each algorithms $A_i$ for LOTUS-Outlier. We use the same search space as MetaOD~\cite{metaod} for a fair comparison.}
    \label{tab:gamaodsearchspace}
\end{table}


\section{Experimental setup}
In the next two subsections, we describe the experimental setup for Automated selection for Outlier detection and Automated selection for clustering. We use leave-one-out strategy for the evaluation of our system, i.e., we take out one dataset at a time from our benchmarks and use only the other datasets in the meta-data. This ensures independent meta-training on all datasets. Our experimental protocol follows standard AutoML practice of comparing against all classifiers as well as an AutoML framework.

\subsection{Outlier Detection Experimental Setup}
For our experiments with Model Selection for Outlier Detection, we use ADBench~\cite{Han2022ADBenchAD} as $\mathcal{D}$ and retrieve all tabular datasets. ADBench is a collection of 46 datasets. We compare LOTUS against MetaOD for outlier detection and 7 other outlier detection algorithms available in PyOD~\cite{Zhao2019PyODAP}, We use the following Baselines: 
\textbf{IForest} (Isolation Forest)~\cite{Liu2008IsolationF}, \textbf{ABOD} (Angle-Based Outlier Detection)~\cite{ABOD},
\textbf{OCSVM} (One-Class Support Vector Machine)~\cite{ocsvm},
\textbf{LODA} (Lightweight Online Detector of Anomalies)~\cite{Pevn2015LodaLO},
\textbf{KNN} (K-Nearest Neighbors)~\cite{knn2},
\textbf{HBOS} (Histogram-Based Outlier Score)~\cite{hbos},
\textbf{LOF} (Local Outlier Factor)~\cite{LOF},
\textbf{COF} (Connectivity-Based Outlier Factor)~\cite{COF}

For experimental consistency, we use the same search space in our experiments as MetaOD to ensure a fair comparison. We use the area under the ROC curve (AUC) as the optimization metric during the search phase.

\subsection{Clustering Experimental Setup}
For our experiments with Model Selection for Clustering algorithms, we use 57 datasets from OpenML~\cite{OpenML2013} which are suitable for clustering. These datasets are selected manually, we aim to select both synthetic and real-world datasets for our setting, to reflect a wide range of problems and improve the generalizability of our results. 
For a robust comparison, we compare our approach with 7 different baselines:
\footnote{The published automated clustering methods mentioned in our related literature section were non-reproducible, despite our efforts and reaching out to the authors. For example, because of missing code, errors while running given code, or missing code and datasets.}
\textbf{Internal Metric optimization}: The first baseline is LOTUS-Clust optimized with Calinski–Harabasz index (CH), we perform a search with CH metric for one hour and then report the mean AMI of the selected pipeline. This baseline uses the same optimization as described in Equation \ref{eq: internalmetricopt}.
\textbf{KMeans}~\cite{kmeans1},
\textbf{OPTICS}~\cite{ankerst1999optics},
\textbf{Affinity Propagation}~\cite{affinityprop},
\textbf{DBSCAN}((Density-Based Spatial Clustering of Applications with Noise))~\cite{dbscan},
\textbf{Mini Batch KMeans}~\cite{minibkmeans},
\textbf{BIRCH}(balanced iterative reducing and clustering using hierarchies). In this scenario, we focus on model selection by requiring clustering algorithms to optimize for external metrics such as Adjusted Mutual Information (AMI) or Adjusted Rand Index (ARI). We recognize that this is just one application of clustering algorithms, and there are various contexts in which the objective extends beyond maximizing an external metric.

\section{Experimental Results}
To comprehensively evaluate the performance of LOTUS against multiple baselines, we employ two complementary and widely accepted evaluation techniques: the Bayesian Wilcoxon signed-rank test (also known as the ROPE test) and Critical Difference (CD) diagrams.

The ROPE test\cite{Ropetutorial} enables a probabilistic comparison between pairs of methods by estimating the likelihood that one method  outperforms another, is worse, or performs equivalently within a predefined threshold. We define the Region of Practical Equivalence (ROPE) as 1\%(as per \citet{Ropetutorial}), which reflects the smallest performance difference considered practically meaningful. While task-specific ROPE values may vary, this threshold offers a reasonable baseline in the absence of domain-specific guidance. We utilize the \texttt{baycomp} library\cite{Ropetutorial} to run and visualize these comparisons.

In parallel, we use Critical Difference(CD) diagrams~\cite{CD} to compare the average rank of each method across all datasets. Unlike the ROPE test, which offers probabilistic insight at the pairwise level and provides information about whether the difference is practically relevant, CD diagrams provide a global view of performance consistency across tasks and help identify statistically significant differences in rankings among multiple algorithms, even if a statistically different rank is not necessarily practically relevant.

Together, these two evaluation strategies paint a comprehensive picture: the ROPE test confirms that LOTUS is statistically more likely to outperform individual alternatives by a practically relevant margin, while the CD diagrams demonstrate that LOTUS consistently achieves top rankings across diverse datasets. This dual validation highlights the robustness and general effectiveness of LOTUS in unsupervised model selection tasks.

\subsection{Outlier detection results}
\begin{figure}[h]
    \centering
    \includegraphics[width=0.70\columnwidth]{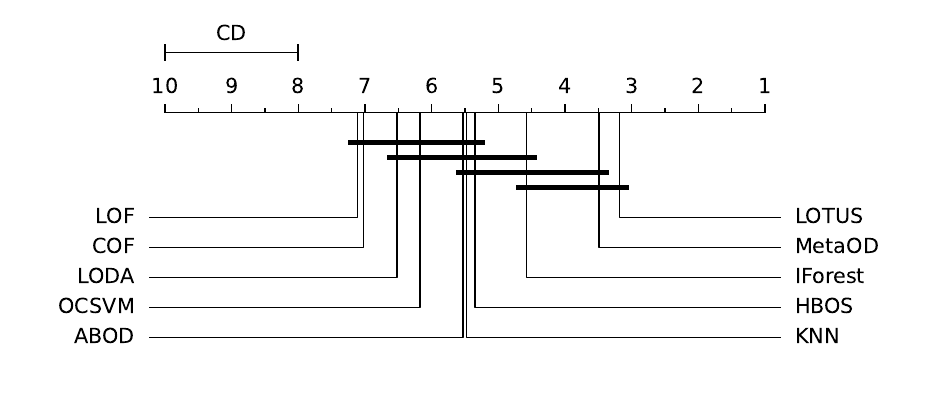}
    \caption{Comparison of average rank
(lower is better) of methods w.r.t. outlier detection performance across datasets in ADBench. The differences in rank of methods connected by horizontal black bars are not a statistically significant.}
    \label{fig:avgrank}
\end{figure}
\begin{figure}[t]
    \centering
    \includegraphics[width=0.5\linewidth]{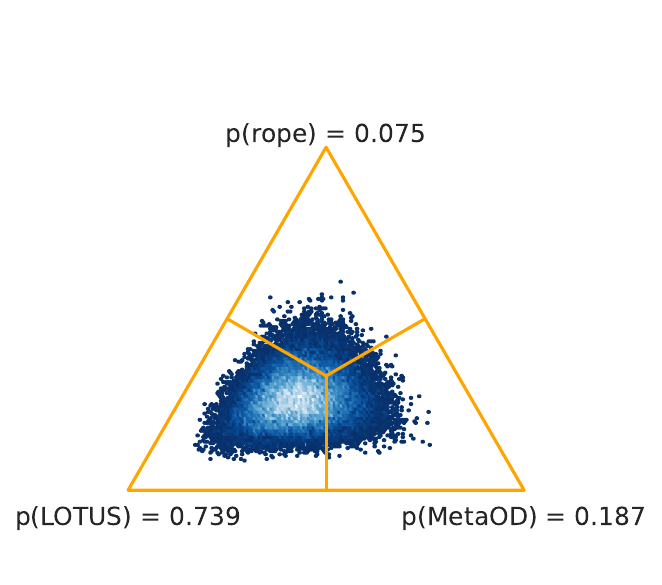}
    \caption{ROPE test result, LOTUS vs MetaOD}
    \label{fig:metaOD}
\end{figure}
\begin{figure*}[]
     \centering
     \begin{subfigure}[b]{0.45\textwidth}
         \centering
         \includegraphics[width=\textwidth]{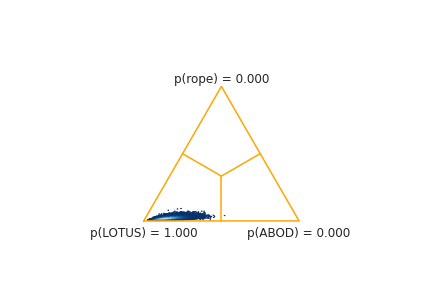}
         \caption{LOTUS vs ABOD}
         \label{fig:y equals x}
     \end{subfigure}
     \hfill
     \begin{subfigure}[b]{0.45\textwidth}
         \centering
         \includegraphics[width=\textwidth]{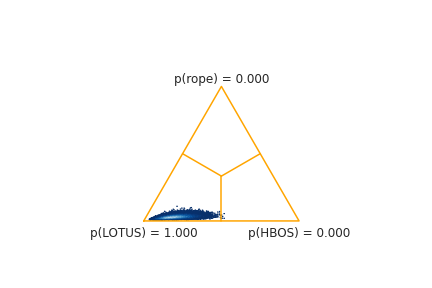}
         \caption{LOTUS vs HBOS}
     \end{subfigure}
     \hfill
     \begin{subfigure}[b]{0.45\textwidth}
         \centering
         \includegraphics[width=\textwidth]{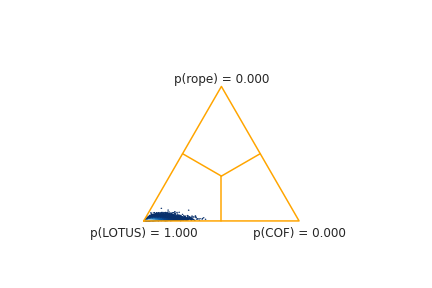}
         \caption{LOTUS vs COF}
         \label{fig:five over x}
     \end{subfigure}
      \centering
     \begin{subfigure}[b]{0.45\textwidth}
         \centering
         \includegraphics[width=\textwidth]{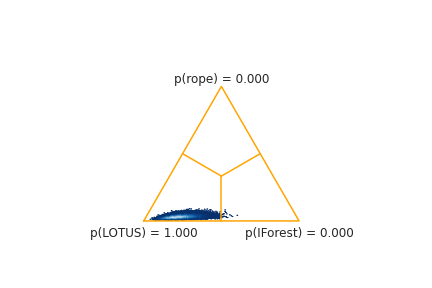}
         \caption{LOTUS vs IForest}
         \label{fig:y equals x}
     \end{subfigure}
     \hfill
     \begin{subfigure}[b]{0.45\textwidth}
         \centering
         \includegraphics[width=\textwidth]{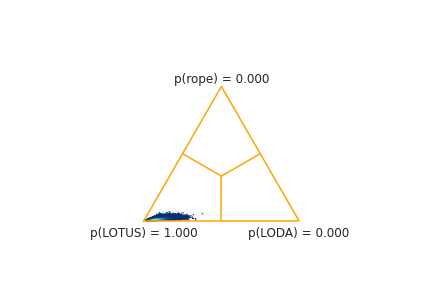}
         \caption{LOTUS vs LODA}
     \end{subfigure}
     \hfill
     \begin{subfigure}[b]{0.45\textwidth}
         \centering
         \includegraphics[width=\textwidth]{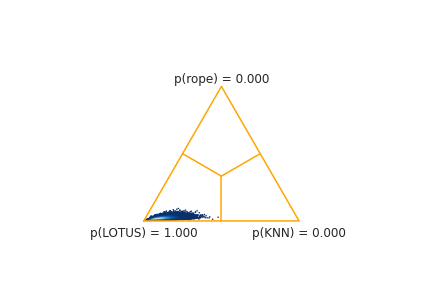}
         \caption{LOTUS vs KNN}
         \label{fig:five over x}
     \end{subfigure}
    \hfill
     \begin{subfigure}[b]{0.45\textwidth}
         \centering
         \includegraphics[width=\textwidth]{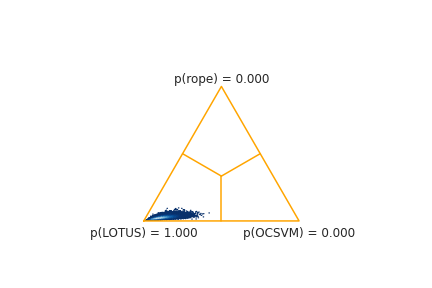}
         \caption{LOTUS vs OCSVM}
         \label{fig:five over x}
     \end{subfigure}
     \begin{subfigure}[b]{0.45\textwidth}
         \centering
         \includegraphics[width=\textwidth]{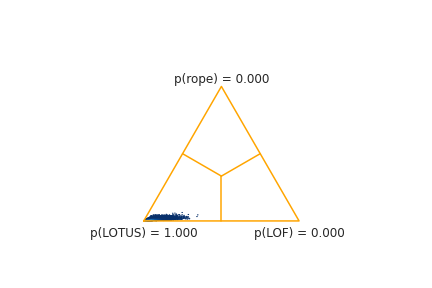}
         \caption{LOTUS vs LOF}
         \label{fig:five over x}
     \end{subfigure}
        \caption{ROPE test result of LOTUS vs (a) ABOD (b) HBOS (c) COF (d) IForest (e) LODA (f) KNN (g) OCSVM (h) LOF}
        \label{fig: all}       
\end{figure*}

The results of the ROPE test comparing LOTUS with individual outlier detection techniques are summarized in Figure \ref{fig: all}. Even when LOTUS is used with its default hyperparameter configuration, LOTUS is better than all baselines ($p(LOTUS) \approx 1 $). We show the pairwise comparison of LOTUS and MetaOD using the ROPE test in Figure \ref{fig:metaOD}. According to ROPE test, for  new datasets from the same distribution as the benchmark datasets, there is a 74\% probability that LOTUS outperforms MetaOD, a 24\% probability the performances are practically equivalent, and a 2\% chance MetaOD is better. This higher performance of LOTUS also shows that LOTUS is more robust than MetaOD.

A complementary global view is provided by the Critical Difference (CD) diagram in Figure \ref{fig:avgrank}. After excluding three datasets on which MetaOD crashed\footnote{MetaOD failed due to invalid parameter settings; the remaining 43 datasets form the basis of the CD plot.}, LOTUS attains the best average rank across all methods, with the only non-significant different ranks being of MetaOD and IForest. Among the classical detectors, Isolation Forest emerges as the strongest single baseline, followed by HBOS and KNN.

Taken together, ROPE probabilities and CD ranking consistently show that LOTUS delivers more accurate and reliable performance on a diverse suite of outlier detection tasks.

\newpage
\begin{figure*}[!ht]
\centering
\begin{subfigure}{0.35\textwidth}
\includegraphics[width=\textwidth]{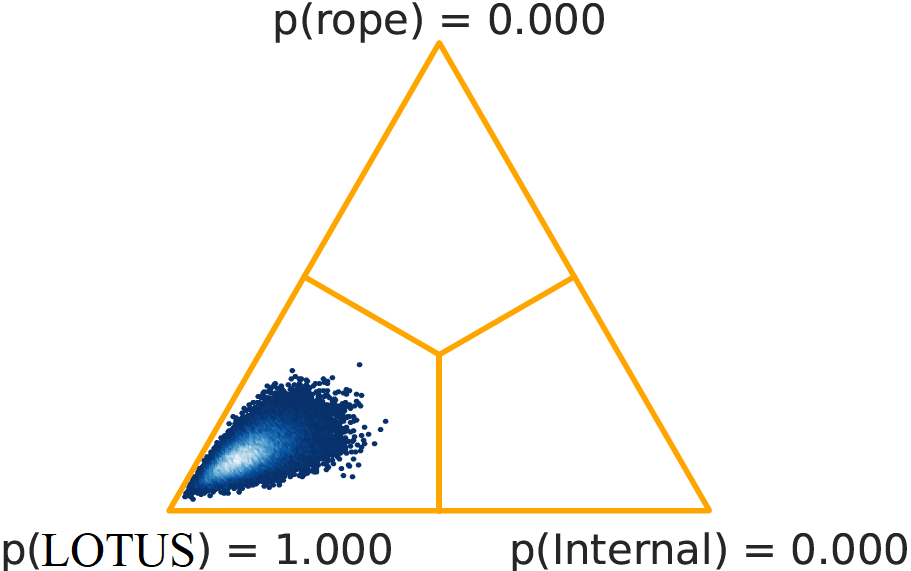}
\caption{}
\end{subfigure}
\begin{subfigure}{0.35\textwidth}
\includegraphics[width=\textwidth]{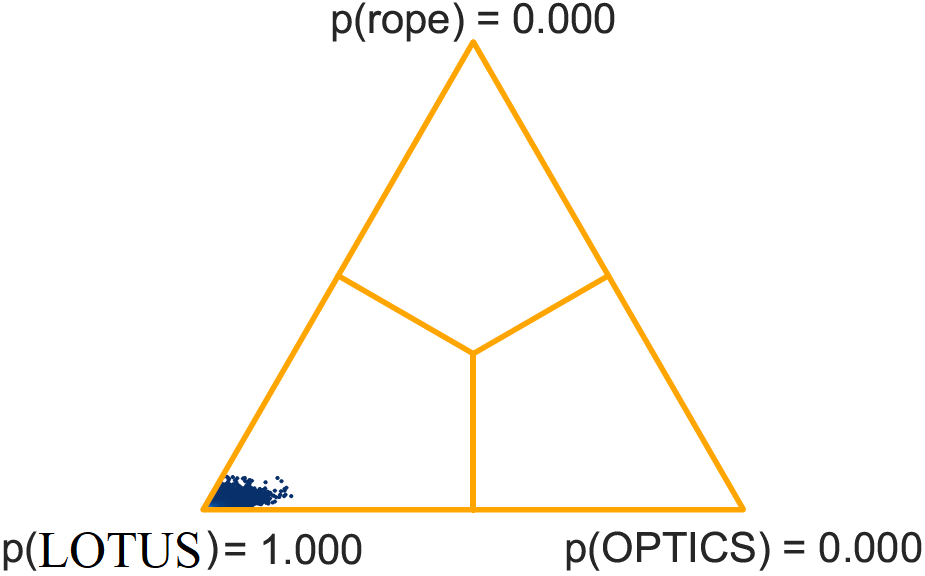}
\caption{}
\end{subfigure}
\begin{subfigure}{0.35\textwidth}
\includegraphics[width=\textwidth]{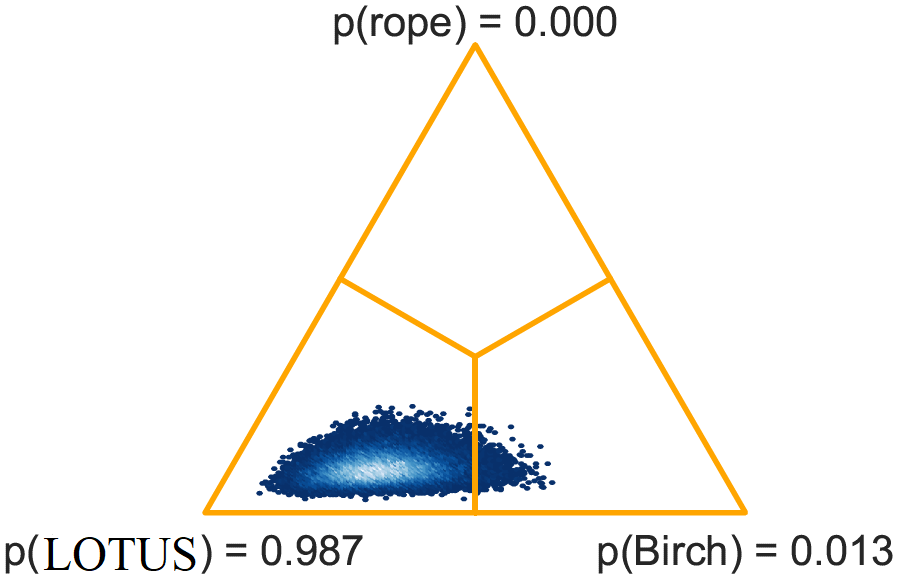}
\caption{}
\end{subfigure}
\begin{subfigure}{0.35\textwidth}
 \includegraphics[width=\textwidth]{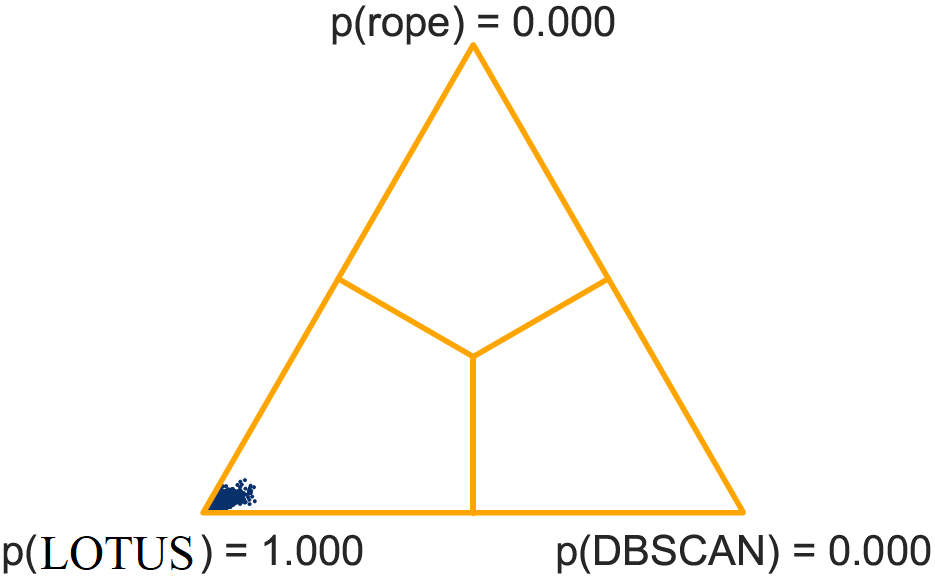}
\caption{}
\end{subfigure}
\begin{subfigure}{0.35\textwidth}
\includegraphics[width=\textwidth]{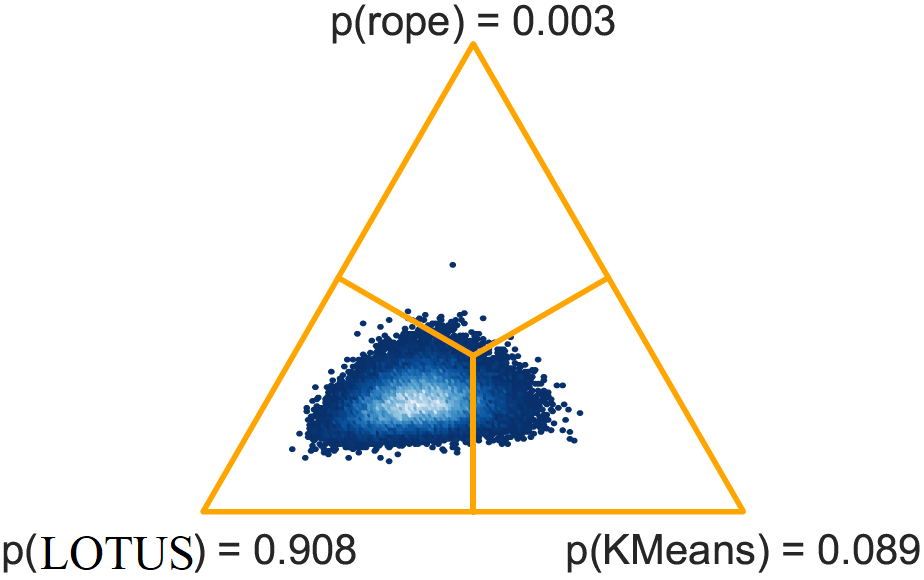}
\caption{}
\end{subfigure}
\begin{subfigure}{0.35\textwidth}
\includegraphics[width=\textwidth]{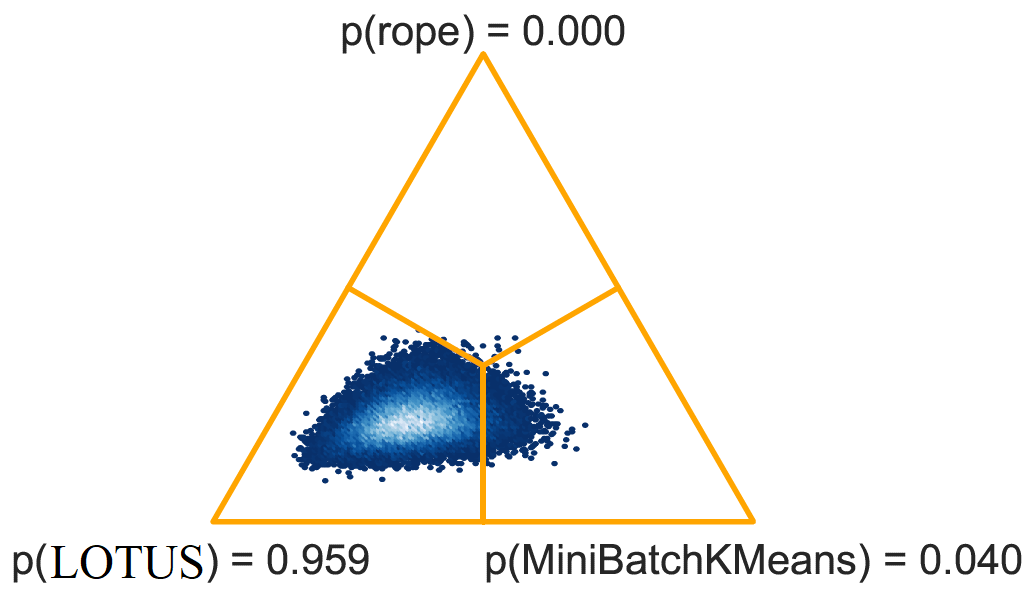}
\caption{}
\end{subfigure}
\begin{subfigure}{0.35\textwidth}
\includegraphics[width=\textwidth]{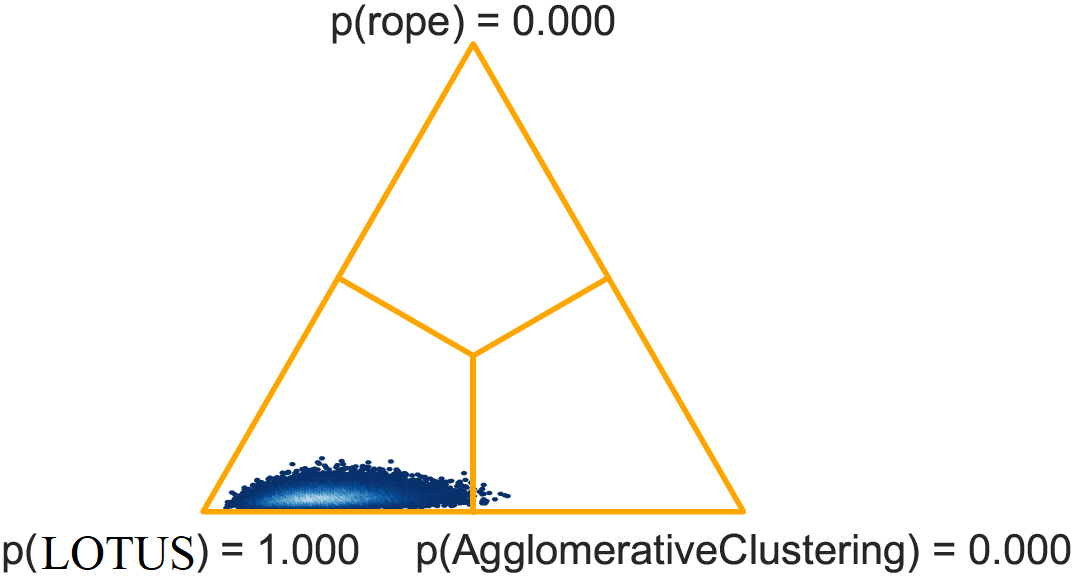}
\caption{}

\end{subfigure}
\caption{Bayesian Wilcoxon signed-rank test results of Our method vs Baselines with ROPE=0.01, this figure shows the simplex and projections of the posterior for the Bayesian sign-rank test. The closer the distribution is to the bottom left corner, the more likely it is that our method is better.}
\label{fig:rope}

\end{figure*}
\subsection{Clustering Results}

To evaluate the effectiveness of LOTUS for an external CVI, we report the average Adjusted Mutual Information (AMI)~\cite{JMLR:metrics} obtained from five runs of both the baselines and our approach. We compare our results against methods which use internal metric optimization as well as standard clustering baselines. Figure \ref{fig:rope} shows the ROPE test results comparing LOTUS to the other methods. We observe that LOTUS is consistently better than Internal metric optimization. This may imply a weak correlation between performance on external and internal CVIs from an AutoML perspective (i.e., an optimized pipeline performing well on an internal CVI for a given task does not imply it performs well on an external metric for the same task). Our algorithm also performed very well against other selected clustering methods. However, in contrast to outlier detection, traditional algorithms like KMeans and MiniBatchKMeans are still very competitive and estimated to outperform even LOTUS with small probability. While LOTUS outperforms other baselines in a setting where one wants to optimize for an external CVI (Like AMI or ARI) without access to labels, we do acknowledge that there are other settings where clustering algorithms may be applied with other objectives.
Additionally, the metrics considered in the evaluation ultimately used ground truth labels, but there are many cases where the clustering dataset does not have ground truth labels.
\par

\begin{figure}[!th]
    \centering
    \includegraphics[width=\textwidth]{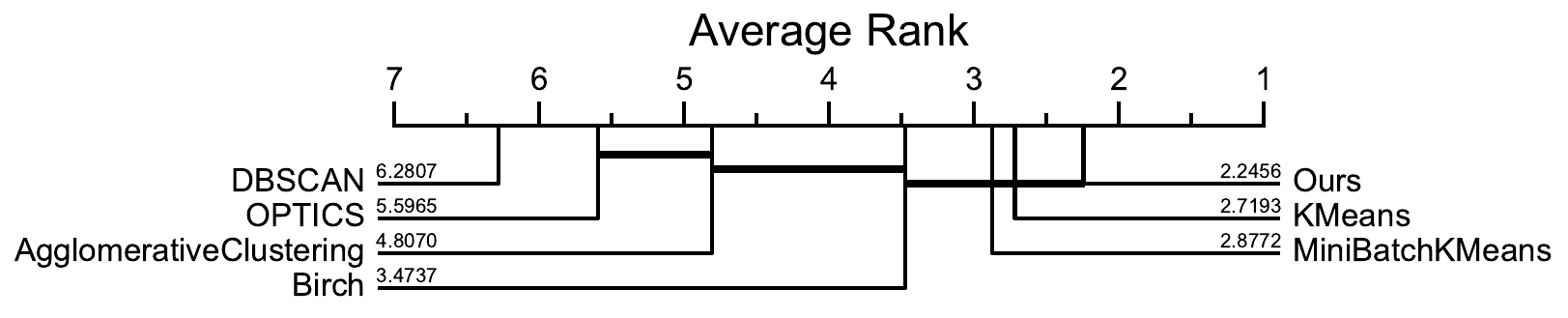}
    \caption{Critical difference diagram of LOTUS vs baselines (Mean performance of all)}
    \label{fig:cd}
\end{figure}

We provide a critical difference diagram\footnote{ We use the code provided here \url{https://github.com/sherbold/autorank } to generate the critical difference diagram for our experiment} in Figure \ref{fig:cd}. LOTUS attains the best average rank across clustering methods (Fig.~\ref{fig:cd}); \texttt{KMeans} and \texttt{MiniBatchKMeans} are the strongest individual baselines. ROPE tests (ROPE=1\%) assign $P(\text{LOTUS})>0.90$ against both, confirming LOTUS is likely to outperform both strong baselines. LOTUS therefore provides more reliable model selection than internal-metric optimization or traditional algorithms.

\section{Conclusion, Discussion, and Limitations}

 In this work, we present LOTUS, a \textit{simple} but very \textit{effective} method for model selection on multiple unsupervised machine learning tasks. We use a unified Gromov-Wasserstein based distance to capture structure dataset similarity. We create a unified framework for model selection for unsupervised machine learning tasks. We show the effectiveness of LOTUS by comparing it empirically with existing baselines and via ROPE test scores show 74\% and 90\% probability of improvement over existing baselines. We believe that our work shows the scope for a unified framework for model selection on unsupervised tasks. By introducing LOTUS we avoid the need for complex meta-feature construction and ensemble of internal and external metric-based optimization to perform AutoML on unsupervised tasks, which greatly simplifies the process of model selection on these tasks. We also introduced two open source AutoML systems to ensure reproducibility and enable adoption and further research by the community.

\subsection{Limitations}
LOTUS used GW distance which proved to be a feasible and robust approach for dataset similarity and meta-learning. We would like to emphasize that this similarity measure should only be used as a relative similarity measure with the objective of finding unsupervised learning algorithms.  For instance, in our case, we use this similarity measure to find the most similar dataset from a collection of datasets in $\mathcal{D}_{meta}$. This highlights an important avenue for future investigation: developing meta-learning techniques within the LOTUS framework that can generalize effectively even from smaller or less directly comparable meta-datasets, potentially through advancements in few-shot learning or more sophisticated transfer learning mechanisms.

In cases where there are no similar datasets our suggested pipeline may not yield favorable results (Though one must note that in our experiments we have not intentionally selected datasets which are similar to each other).  In cases  where there are no similar datasets, such as with dataset id \textit{42464} in our clustering experiments, our suggested pipeline did not yield favorable results. Second, the time complexity of our system scales linearly with the number of datasets in $\mathcal{D}_{meta}$. This means that as the number of datasets increases, LOTUS may require more time to perform model selection. Though these limitations are important to note as they may impact the practical application of our approach in certain scenarios, LOTUS does still provide a working pipeline for an unsupervised task where the tuning might not even be possible because of a lack of ground truth. To overcome this limitation, we make our meta-dataset public and will keep adding more datasets and optimal models there.

\subsection{Future Work}
We believe that there can be two promising future work directions with LOTUS:
\begin{enumerate}
    \item
The optimization for similarity calculation can be done in more efficient ways. ICNN~\cite{icnn} can be used to compute GW distance for faster computation, although these networks do not (yet) support Gromov Wasserstein space. We believe that faster approximations of LOTUS enables real-time pipeline recommendations, which is critical in dynamic or resource-constrained environments.

\item LOTUS can be easily extended to other unsupervised tasks like unsupervised time series outlier detection, online clustering and time series clustering. Extending LOTUS to these diverse domains would further underscore its potential as a generalizable meta-learning architecture for a wide spectrum of unsupervised AutoML challenges.
\end{enumerate}

\bibliography{main}


\begin{thebibliography}{57}
\ifx \bisbn   \undefined \def \bisbn  #1{ISBN #1}\fi
\ifx \binits  \undefined \def \binits#1{#1}\fi
\ifx \bauthor  \undefined \def \bauthor#1{#1}\fi
\ifx \batitle  \undefined \def \batitle#1{#1}\fi
\ifx \bjtitle  \undefined \def \bjtitle#1{#1}\fi
\ifx \bvolume  \undefined \def \bvolume#1{\textbf{#1}}\fi
\ifx \byear  \undefined \def \byear#1{#1}\fi
\ifx \bissue  \undefined \def \bissue#1{#1}\fi
\ifx \bfpage  \undefined \def \bfpage#1{#1}\fi
\ifx \blpage  \undefined \def \blpage #1{#1}\fi
\ifx \burl  \undefined \def \burl#1{\textsf{#1}}\fi
\ifx \doiurl  \undefined \def \doiurl#1{\url{https://doi.org/#1}}\fi
\ifx \betal  \undefined \def \betal{\textit{et al.}}\fi
\ifx \binstitute  \undefined \def \binstitute#1{#1}\fi
\ifx \binstitutionaled  \undefined \def \binstitutionaled#1{#1}\fi
\ifx \bctitle  \undefined \def \bctitle#1{#1}\fi
\ifx \beditor  \undefined \def \beditor#1{#1}\fi
\ifx \bpublisher  \undefined \def \bpublisher#1{#1}\fi
\ifx \bbtitle  \undefined \def \bbtitle#1{#1}\fi
\ifx \bedition  \undefined \def \bedition#1{#1}\fi
\ifx \bseriesno  \undefined \def \bseriesno#1{#1}\fi
\ifx \blocation  \undefined \def \blocation#1{#1}\fi
\ifx \bsertitle  \undefined \def \bsertitle#1{#1}\fi
\ifx \bsnm \undefined \def \bsnm#1{#1}\fi
\ifx \bsuffix \undefined \def \bsuffix#1{#1}\fi
\ifx \bparticle \undefined \def \bparticle#1{#1}\fi
\ifx \barticle \undefined \def \barticle#1{#1}\fi
\bibcommenthead
\ifx \bconfdate \undefined \def \bconfdate #1{#1}\fi
\ifx \botherref \undefined \def \botherref #1{#1}\fi
\ifx \url \undefined \def \url#1{\textsf{#1}}\fi
\ifx \bchapter \undefined \def \bchapter#1{#1}\fi
\ifx \bbook \undefined \def \bbook#1{#1}\fi
\ifx \bcomment \undefined \def \bcomment#1{#1}\fi
\ifx \oauthor \undefined \def \oauthor#1{#1}\fi
\ifx \citeauthoryear \undefined \def \citeauthoryear#1{#1}\fi
\ifx \endbibitem  \undefined \def \endbibitem {}\fi
\ifx \bconflocation  \undefined \def \bconflocation#1{#1}\fi
\ifx \arxivurl  \undefined \def \arxivurl#1{\textsf{#1}}\fi
\csname PreBibitemsHook\endcsname

\bibitem[\protect\citeauthoryear{Hutter et~al.}{2019}]{Hutter2019AutomatedML}
\begin{botherref}
\oauthor{\bsnm{Hutter}, \binits{F.}},
\oauthor{\bsnm{Kotthoff}, \binits{L.}},
\oauthor{\bsnm{Vanschoren}, \binits{J.}}:
Automated machine learning: Methods, systems, challenges.
Automated Machine Learning
(2019)
\end{botherref}
\endbibitem

\bibitem[\protect\citeauthoryear{Shchur et~al.}{2023}]{Shchur2023AutoGluonTimeSeriesAF}
\begin{botherref}
\oauthor{\bsnm{Shchur}, \binits{O.}},
\oauthor{\bsnm{Turkmen}, \binits{C.}},
\oauthor{\bsnm{Erickson}, \binits{N.}},
\oauthor{\bsnm{Shen}, \binits{H.}},
\oauthor{\bsnm{Shirkov}, \binits{A.}},
\oauthor{\bsnm{Hu}, \binits{T.}},
\oauthor{\bsnm{Wang}, \binits{B.}}:
Autogluon-timeseries: Automl for probabilistic time series forecasting.
AutoML Conference
(2023)
\end{botherref}
\endbibitem

\bibitem[\protect\citeauthoryear{Feurer et~al.}{2015}]{NIPS2015_11d0e628}
\begin{bchapter}
\bauthor{\bsnm{Feurer}, \binits{M.}},
\bauthor{\bsnm{Klein}, \binits{A.}},
\bauthor{\bsnm{Eggensperger}, \binits{K.}},
\bauthor{\bsnm{Springenberg}, \binits{J.}},
\bauthor{\bsnm{Blum}, \binits{M.}},
\bauthor{\bsnm{Hutter}, \binits{F.}}:
\bctitle{Efficient and robust automated machine learning}.
In: \bbtitle{Advances in Neural Information Processing Systems}
(\byear{2015})
\end{bchapter}
\endbibitem

\bibitem[\protect\citeauthoryear{Gijsbers and Vanschoren}{2021}]{gama}
\begin{bchapter}
\bauthor{\bsnm{Gijsbers}, \binits{P.}},
\bauthor{\bsnm{Vanschoren}, \binits{J.}}:
\bctitle{Gama: A general automated machine learning assistant}.
In: \bbtitle{Machine Learning and Knowledge Discovery in Databases. Applied Data Science and Demo Track}
(\byear{2021})
\end{bchapter}
\endbibitem

\bibitem[\protect\citeauthoryear{Vanschoren}{2018}]{Vanschoren2018MetaLearningAS}
\begin{botherref}
\oauthor{\bsnm{Vanschoren}, \binits{J.}}:
Meta-learning: A survey.
ArXiv
\textbf{abs/1810.03548}
(2018)
\end{botherref}
\endbibitem

\bibitem[\protect\citeauthoryear{Feurer et~al.}{2022}]{feurer-arxiv20a}
\begin{botherref}
\oauthor{\bsnm{Feurer}, \binits{M.}},
\oauthor{\bsnm{Eggensperger}, \binits{K.}},
\oauthor{\bsnm{Falkner}, \binits{S.}},
\oauthor{\bsnm{Lindauer}, \binits{M.}},
\oauthor{\bsnm{Hutter}, \binits{F.}}:
Auto-sklearn 2.0: hands-free automl via meta-learning.
J. Mach. Learn. Res.
\textbf{23}(1)
(2022)
\end{botherref}
\endbibitem

\bibitem[\protect\citeauthoryear{Wang et~al.}{2021}]{Wang2021FLAMLAF}
\begin{bchapter}
\bauthor{\bsnm{Wang}, \binits{C.}},
\bauthor{\bsnm{Wu}, \binits{Q.}},
\bauthor{\bsnm{Weimer}, \binits{M.}},
\bauthor{\bsnm{Zhu}, \binits{E.}}:
\bctitle{Flaml: A fast and lightweight automl library}.
In: \bbtitle{MLSys}
(\byear{2021})
\end{bchapter}
\endbibitem

\bibitem[\protect\citeauthoryear{Zhao et~al.}{}]{metaod}
\begin{botherref}
\oauthor{\bsnm{Zhao}, \binits{Y.}},
\oauthor{\bsnm{Rossi}, \binits{R.}},
\oauthor{\bsnm{Akoglu}, \binits{L.}}:
Automatic unsupervised outlier model selection.
In: Ranzato, M., Beygelzimer, A., Dauphin, Y., Liang, P.S., Vaughan, J.W. (eds.)
Advances in Neural Information Processing Systems
\end{botherref}
\endbibitem

\bibitem[\protect\citeauthoryear{Anonymous}{2024}]{anonymous2024hpod}
\begin{bchapter}
\bauthor{\bsnm{Anonymous}}:
\bctitle{{HPOD}: Hyperparameter optimization for unsupervised outlier detection}.
In: \bbtitle{AutoML 2024 Methods Track}
(\byear{2024})
\end{bchapter}
\endbibitem

\bibitem[\protect\citeauthoryear{Vu et~al.}{}]{TautoOD}
\begin{botherref}
\oauthor{\bsnm{Vu}, \binits{L.}},
\oauthor{\bsnm{Kirchner}, \binits{P.}},
\oauthor{\bsnm{Aggarwal}, \binits{C.C.}},
\oauthor{\bsnm{Samulowitz}, \binits{H.}}:
Instance-level metalearning for outlier detection.
In: Proceedings of the Thirty-Third International Joint Conference on Artificial Intelligence, {IJCAI-24}
\end{botherref}
\endbibitem

\bibitem[\protect\citeauthoryear{Stern et~al.}{2010}]{stern2010collaborative}
\begin{bchapter}
\bauthor{\bsnm{Stern}, \binits{D.}},
\bauthor{\bsnm{Herbrich}, \binits{R.}},
\bauthor{\bsnm{Graepel}, \binits{T.}},
\bauthor{\bsnm{Samulowitz}, \binits{H.}},
\bauthor{\bsnm{Pulina}, \binits{L.}},
\bauthor{\bsnm{Tacchella}, \binits{A.}}:
\bctitle{Collaborative expert portfolio management}.
In: \bbtitle{Proceedings of the Twenty-Fourth AAAI Conference on Artificial Intelligence AAAI-10 (to Appear)}
(\byear{2010})
\end{bchapter}
\endbibitem

\bibitem[\protect\citeauthoryear{Li et~al.}{2020}]{Li2020PyODDSAE}
\begin{botherref}
\oauthor{\bsnm{Li}, \binits{Y.}},
\oauthor{\bsnm{Zha}, \binits{D.}},
\oauthor{\bsnm{Zou}, \binits{N.}},
\oauthor{\bsnm{Hu}, \binits{X.}}:
Pyodds: An end-to-end outlier detection system with automated machine learning.
Companion Proceedings of the Web Conference 2020
(2020)
\end{botherref}
\endbibitem

\bibitem[\protect\citeauthoryear{Goix}{2016}]{emmv}
\begin{botherref}
\oauthor{\bsnm{Goix}, \binits{N.}}:
How to Evaluate the Quality of Unsupervised Anomaly Detection Algorithms?
arXiv
(2016).
\doiurl{10.48550/ARXIV.1607.01152} .
\url{https://arxiv.org/abs/1607.01152}
\end{botherref}
\endbibitem

\bibitem[\protect\citeauthoryear{Marques et~al.}{2015}]{IREOS}
\begin{bchapter}
\bauthor{\bsnm{Marques}, \binits{H.O.}},
\bauthor{\bsnm{Campello}, \binits{R.J.G.B.}},
\bauthor{\bsnm{Zimek}, \binits{A.}},
\bauthor{\bsnm{Sander}, \binits{J.}}:
\bctitle{On the internal evaluation of unsupervised outlier detection}.
In: \bbtitle{Proceedings of the 27th International Conference on Scientific and Statistical Database Management}
(\byear{2015})
\end{bchapter}
\endbibitem

\bibitem[\protect\citeauthoryear{Ma et~al.}{2021}]{internalstrategysurvey}
\begin{botherref}
\oauthor{\bsnm{Ma}, \binits{M.Q.}},
\oauthor{\bsnm{Zhao}, \binits{Y.}},
\oauthor{\bsnm{Zhang}, \binits{X.}},
\oauthor{\bsnm{Akoglu}, \binits{L.}}:
A large-scale study on unsupervised outlier model selection: Do internal strategies suffice?
CoRR
\textbf{abs/2104.01422}
(2021)
{\href{https://arxiv.org/abs/2104.01422}{{2104.01422}}}
\end{botherref}
\endbibitem

\bibitem[\protect\citeauthoryear{Treder-Tschechlov et~al.}{2023}]{ML2DAC}
\begin{botherref}
\oauthor{\bsnm{Treder-Tschechlov}, \binits{D.}},
\oauthor{\bsnm{Fritz}, \binits{M.}},
\oauthor{\bsnm{Schwarz}, \binits{H.}},
\oauthor{\bsnm{Mitschang}, \binits{B.}}:
Ml2dac: Meta-learning to democratize automl for clustering analysis.
Proc. ACM Manag. Data
\textbf{1}(2)
(2023)
\doiurl{10.1145/3589289}
\end{botherref}
\endbibitem

\bibitem[\protect\citeauthoryear{Liu et~al.}{}]{autocluster}
\begin{botherref}
\oauthor{\bsnm{Liu}, \binits{Y.}},
\oauthor{\bsnm{Li}, \binits{S.}},
\oauthor{\bsnm{Tian}, \binits{W.}}:
Autocluster: Meta-learning based ensemble method for automated unsupervised clustering.
In: Advances in Knowledge Discovery and Data Mining: 25th Pacific-Asia Conference, PAKDD 2021,
pp. 246--258.
Springer,
Berlin, Heidelberg
\end{botherref}
\endbibitem

\bibitem[\protect\citeauthoryear{Liao et~al.}{2016}]{clusteringapp}
\begin{botherref}
\oauthor{\bsnm{Liao}, \binits{M.}},
\oauthor{\bsnm{Li}, \binits{Y.}},
\oauthor{\bsnm{Kianifard}, \binits{F.}},
\oauthor{\bsnm{Obi}, \binits{E.N.}},
\oauthor{\bsnm{Arcona}, \binits{S.}}:
Cluster analysis and its application to healthcare claims data: a study of end-stage renal disease patients who initiated hemodialysis.
BMC Nephrology
\textbf{17}
(2016)
\end{botherref}
\endbibitem

\bibitem[\protect\citeauthoryear{Liu et~al.}{}]{liu2021autocluster}
\begin{botherref}
\oauthor{\bsnm{Liu}, \binits{Y.}},
\oauthor{\bsnm{Li}, \binits{S.}},
\oauthor{\bsnm{Tian}, \binits{W.}}:
Autocluster: Meta-learning based ensemble method for automated unsupervised clustering.
In: Pacific-Asia Conference on Knowledge Discovery and Data Mining
\end{botherref}
\endbibitem

\bibitem[\protect\citeauthoryear{Poulakis et~al.}{2020}]{poulakis2020autoclust}
\begin{bchapter}
\bauthor{\bsnm{Poulakis}, \binits{Y.}},
\bauthor{\bsnm{Doulkeridis}, \binits{C.}},
\bauthor{\bsnm{Kyriazis}, \binits{D.}}:
\bctitle{Autoclust: A framework for automated clustering based on cluster validity indices}.
In: \bbtitle{2020 IEEE International Conference on Data Mining (ICDM)},
pp. \bfpage{1220}--\blpage{1225}
(\byear{2020}).
\bcomment{IEEE}
\end{bchapter}
\endbibitem

\bibitem[\protect\citeauthoryear{Tschechlov et~al.}{2021}]{tschechlov2021automl4clust}
\begin{bchapter}
\bauthor{\bsnm{Tschechlov}, \binits{D.}},
\bauthor{\bsnm{Fritz}, \binits{M.}},
\bauthor{\bsnm{Schwarz}, \binits{H.}},
\bauthor{\bsnm{Velegrakis}, \binits{Y.}},
\bauthor{\bsnm{Zeinalipour-Yazti}, \binits{D.}},
\bauthor{\bsnm{Chrysanthis}, \binits{P.}},
\bauthor{\bsnm{Guerra}, \binits{F.}}:
\bctitle{Automl4clust: Efficient automl for clustering analyses.}
In: \bbtitle{EDBT},
pp. \bfpage{343}--\blpage{348}
(\byear{2021})
\end{bchapter}
\endbibitem

\bibitem[\protect\citeauthoryear{Singh and Vanschoren}{}]{ijcai2023p843}
\begin{botherref}
\oauthor{\bsnm{Singh}, \binits{P.}},
\oauthor{\bsnm{Vanschoren}, \binits{J.}}:
Automl for outlier detection with optimal transport distances.
In: Proceedings of the Thirty-Second International Joint Conference on Artificial Intelligence, {IJCAI-23}.
Demo Track
\end{botherref}
\endbibitem

\bibitem[\protect\citeauthoryear{Villani}{2008}]{Villani2008OptimalTO}
\begin{bchapter}
\bauthor{\bsnm{Villani}, \binits{C.}}:
\bctitle{Optimal transport: Old and new}.
In: \bbtitle{Optimal Transport}
(\byear{2008})
\end{bchapter}
\endbibitem

\bibitem[\protect\citeauthoryear{Cuturi}{2013}]{Cuturi2013SinkhornDL}
\begin{bchapter}
\bauthor{\bsnm{Cuturi}, \binits{M.}}:
\bctitle{Sinkhorn distances: Lightspeed computation of optimal transport}.
In: \bbtitle{NIPS}
(\byear{2013})
\end{bchapter}
\endbibitem

\bibitem[\protect\citeauthoryear{M{\'e}moli}{2011}]{Mmoli2011GromovWassersteinDA}
\begin{barticle}
\bauthor{\bsnm{M{\'e}moli}, \binits{F.}}:
\batitle{Gromov–wasserstein distances and the metric approach to object matching}.
\bjtitle{Foundations of Computational Mathematics}
\bvolume{11},
\bfpage{417}--\blpage{487}
(\byear{2011})
\end{barticle}
\endbibitem

\bibitem[\protect\citeauthoryear{Mémoli et~al.}{2018}]{mémoli2018sketchingclusteringmetricmeasure}
\begin{botherref}
\oauthor{\bsnm{Mémoli}, \binits{F.}},
\oauthor{\bsnm{Sidiropoulos}, \binits{A.}},
\oauthor{\bsnm{Singhal}, \binits{K.}}:
Sketching and Clustering Metric Measure Spaces
(2018).
\url{https://arxiv.org/abs/1801.00551}
\end{botherref}
\endbibitem

\bibitem[\protect\citeauthoryear{Scetbon et~al.}{}]{gwlr}
\begin{botherref}
\oauthor{\bsnm{Scetbon}, \binits{M.}},
\oauthor{\bsnm{Peyr{\'e}}, \binits{G.}},
\oauthor{\bsnm{Cuturi}, \binits{M.}}:
Linear-time gromov {W}asserstein distances using low rank couplings and costs.
In: Proceedings of the 39th International Conference on Machine Learning
\end{botherref}
\endbibitem

\bibitem[\protect\citeauthoryear{Peyré et~al.}{}]{pmlr-v48-peyre16}
\begin{botherref}
\oauthor{\bsnm{Peyré}, \binits{G.}},
\oauthor{\bsnm{Cuturi}, \binits{M.}},
\oauthor{\bsnm{Solomon}, \binits{J.}}:
Gromov-wasserstein averaging of kernel and distance matrices.
In: Proceedings of The 33rd International Conference on Machine Learning
\end{botherref}
\endbibitem

\bibitem[\protect\citeauthoryear{Scetbon et~al.}{}]{pmlr-v139-scetbon21a}
\begin{botherref}
\oauthor{\bsnm{Scetbon}, \binits{M.}},
\oauthor{\bsnm{Cuturi}, \binits{M.}},
\oauthor{\bsnm{Peyr{\'e}}, \binits{G.}}:
Low-rank sinkhorn factorization.
In: Proceedings of the 38th International Conference on Machine Learning
\end{botherref}
\endbibitem

\bibitem[\protect\citeauthoryear{Scetbon and Cuturi}{2022}]{Scetbon2022LowrankOT}
\begin{botherref}
\oauthor{\bsnm{Scetbon}, \binits{M.}},
\oauthor{\bsnm{Cuturi}, \binits{M.}}:
Low-rank optimal transport: Approximation, statistics and debiasing.
NeurIPS 2022
\textbf{abs/2205.12365}
(2022)
\end{botherref}
\endbibitem

\bibitem[\protect\citeauthoryear{Hyv{\"a}rinen and Oja}{2000}]{Hyvrinen2000IndependentCA}
\begin{barticle}
\bauthor{\bsnm{Hyv{\"a}rinen}, \binits{A.}},
\bauthor{\bsnm{Oja}, \binits{E.}}:
\batitle{Independent component analysis: algorithms and applications}.
\bjtitle{Neural networks : the official journal of the International Neural Network Society}
\bvolume{13 4-5},
\bfpage{411}--\blpage{30}
(\byear{2000})
\end{barticle}
\endbibitem

\bibitem[\protect\citeauthoryear{Gijsbers and Vanschoren}{2021}]{Gijsbers2021}
\begin{barticle}
\bauthor{\bsnm{Gijsbers}, \binits{P.}},
\bauthor{\bsnm{Vanschoren}, \binits{J.}}:
\batitle{Gama: A general automated machine learning assistant}.
\bjtitle{Lecture Notes in Computer Science (including subseries Lecture Notes in Artificial Intelligence and Lecture Notes in Bioinformatics)}
\bvolume{12461 LNAI},
\bfpage{560}--\blpage{564}
(\byear{2021})
\doiurl{10.1007/978-3-030-67670-4_39}
\end{barticle}
\endbibitem

\bibitem[\protect\citeauthoryear{Li et~al.}{2020}]{asha}
\begin{bchapter}
\bauthor{\bsnm{Li}, \binits{L.}},
\bauthor{\bsnm{Jamieson}, \binits{K.}},
\bauthor{\bsnm{Rostamizadeh}, \binits{A.}},
\bauthor{\bsnm{Gonina}, \binits{E.}},
\bauthor{\bsnm{Ben-tzur}, \binits{J.}},
\bauthor{\bsnm{Hardt}, \binits{M.}},
\bauthor{\bsnm{Recht}, \binits{B.}},
\bauthor{\bsnm{Talwalkar}, \binits{A.}}:
\bctitle{A system for massively parallel hyperparameter tuning}.
In: \beditor{\bsnm{Dhillon}, \binits{I.}},
\beditor{\bsnm{Papailiopoulos}, \binits{D.}},
\beditor{\bsnm{Sze}, \binits{V.}} (eds.)
\bbtitle{Proceedings of Machine Learning and Systems},
pp. \bfpage{230}--\blpage{246}
(\byear{2020})
\end{bchapter}
\endbibitem

\bibitem[\protect\citeauthoryear{Vinh et~al.}{2010}]{AMI}
\begin{barticle}
\bauthor{\bsnm{Vinh}, \binits{N.X.}},
\bauthor{\bsnm{Epps}, \binits{J.}},
\bauthor{\bsnm{Bailey}, \binits{J.}}:
\batitle{Information theoretic measures for clusterings comparison: Variants, properties, normalization and correction for chance}.
\bjtitle{Journal of Machine Learning Research}
\bvolume{11}(\bissue{95}),
\bfpage{2837}--\blpage{2854}
(\byear{2010})
\end{barticle}
\endbibitem

\bibitem[\protect\citeauthoryear{Cali{\'n}ski and Harabasz}{1974}]{calinski1974dendrite}
\begin{barticle}
\bauthor{\bsnm{Cali{\'n}ski}, \binits{T.}},
\bauthor{\bsnm{Harabasz}, \binits{J.}}:
\batitle{A dendrite method for cluster analysis}.
\bjtitle{Communications in Statistics-theory and Methods}
\bvolume{3}(\bissue{1}),
\bfpage{1}--\blpage{27}
(\byear{1974})
\end{barticle}
\endbibitem

\bibitem[\protect\citeauthoryear{Pedregosa et~al.}{2011}]{scikit-learn}
\begin{barticle}
\bauthor{\bsnm{Pedregosa}, \binits{F.}},
\bauthor{\bsnm{Varoquaux}, \binits{G.}},
\bauthor{\bsnm{Gramfort}, \binits{A.}},
\bauthor{\bsnm{Michel}, \binits{V.}},
\bauthor{\bsnm{Thirion}, \binits{B.}},
\bauthor{\bsnm{Grisel}, \binits{O.}},
\bauthor{\bsnm{Blondel}, \binits{M.}},
\bauthor{\bsnm{Prettenhofer}, \binits{P.}},
\bauthor{\bsnm{Weiss}, \binits{R.}},
\bauthor{\bsnm{Dubourg}, \binits{V.}},
\bauthor{\bsnm{Vanderplas}, \binits{J.}},
\bauthor{\bsnm{Passos}, \binits{A.}},
\bauthor{\bsnm{Cournapeau}, \binits{D.}},
\bauthor{\bsnm{Brucher}, \binits{M.}},
\bauthor{\bsnm{Perrot}, \binits{M.}},
\bauthor{\bsnm{Duchesnay}, \binits{E.}}:
\batitle{Scikit-learn: Machine learning in {P}ython}.
\bjtitle{Journal of Machine Learning Research}
\bvolume{12},
\bfpage{2825}--\blpage{2830}
(\byear{2011})
\end{barticle}
\endbibitem

\bibitem[\protect\citeauthoryear{Han et~al.}{2022}]{Han2022ADBenchAD}
\begin{bchapter}
\bauthor{\bsnm{Han}, \binits{S.}},
\bauthor{\bsnm{Hu}, \binits{X.}},
\bauthor{\bsnm{Huang}, \binits{H.}},
\bauthor{\bsnm{Jiang}, \binits{M.}},
\bauthor{\bsnm{Zhao}, \binits{Y.}}:
\bctitle{{ADB}ench: Anomaly detection benchmark}.
In: \bbtitle{Thirty-sixth Conference on Neural Information Processing Systems Datasets and Benchmarks Track}
(\byear{2022})
\end{bchapter}
\endbibitem

\bibitem[\protect\citeauthoryear{Zhao et~al.}{2019}]{Zhao2019PyODAP}
\begin{barticle}
\bauthor{\bsnm{Zhao}, \binits{Y.}},
\bauthor{\bsnm{Nasrullah}, \binits{Z.}},
\bauthor{\bsnm{Li}, \binits{Z.}}:
\batitle{Pyod: A python toolbox for scalable outlier detection}.
\bjtitle{J. Mach. Learn. Res.}
\bvolume{20},
\bfpage{96}--\blpage{1967}
(\byear{2019})
\end{barticle}
\endbibitem

\bibitem[\protect\citeauthoryear{Liu et~al.}{2008}]{Liu2008IsolationF}
\begin{botherref}
\oauthor{\bsnm{Liu}, \binits{F.T.}},
\oauthor{\bsnm{Ting}, \binits{K.M.}},
\oauthor{\bsnm{Zhou}, \binits{Z.-H.}}:
Isolation forest.
2008 Eighth IEEE International Conference on Data Mining,
413--422
(2008)
\end{botherref}
\endbibitem

\bibitem[\protect\citeauthoryear{Kriegel et~al.}{2008}]{ABOD}
\begin{bchapter}
\bauthor{\bsnm{Kriegel}, \binits{H.-P.}},
\bauthor{\bsnm{Schubert}, \binits{M.}},
\bauthor{\bsnm{Zimek}, \binits{A.}}:
\bctitle{Angle-based outlier detection in high-dimensional data}.
In: \bbtitle{Proceedings of the 14th ACM SIGKDD International Conference on Knowledge Discovery and Data Mining},
pp. \bfpage{444}--\blpage{452}
(\byear{2008}).
\doiurl{10.1145/1401890.1401946}
\end{bchapter}
\endbibitem

\bibitem[\protect\citeauthoryear{Sch{\"o}lkopf et~al.}{1999}]{ocsvm}
\begin{bchapter}
\bauthor{\bsnm{Sch{\"o}lkopf}, \binits{B.}},
\bauthor{\bsnm{Williamson}, \binits{R.C.}},
\bauthor{\bsnm{Smola}, \binits{A.}},
\bauthor{\bsnm{Shawe-Taylor}, \binits{J.}},
\bauthor{\bsnm{Platt}, \binits{J.C.}}:
\bctitle{Support vector method for novelty detection}.
In: \bbtitle{NIPS}
(\byear{1999})
\end{bchapter}
\endbibitem

\bibitem[\protect\citeauthoryear{Pevn{\'y}}{2015}]{Pevn2015LodaLO}
\begin{barticle}
\bauthor{\bsnm{Pevn{\'y}}, \binits{T.}}:
\batitle{Loda: Lightweight on-line detector of anomalies}.
\bjtitle{Machine Learning}
\bvolume{102},
\bfpage{275}--\blpage{304}
(\byear{2015})
\end{barticle}
\endbibitem

\bibitem[\protect\citeauthoryear{Angiulli and Pizzuti}{2002}]{knn2}
\begin{bchapter}
\bauthor{\bsnm{Angiulli}, \binits{F.}},
\bauthor{\bsnm{Pizzuti}, \binits{C.}}:
\bctitle{Fast outlier detection in high dimensional spaces}.
In: \bbtitle{PKDD}
(\byear{2002})
\end{bchapter}
\endbibitem

\bibitem[\protect\citeauthoryear{Goldstein and Dengel}{2012}]{hbos}
\begin{bchapter}
\bauthor{\bsnm{Goldstein}, \binits{M.}},
\bauthor{\bsnm{Dengel}, \binits{A.R.}}:
\bctitle{Histogram-based outlier score (hbos): A fast unsupervised anomaly detection algorithm}.
(\byear{2012})
\end{bchapter}
\endbibitem

\bibitem[\protect\citeauthoryear{Breunig et~al.}{2000}]{LOF}
\begin{barticle}
\bauthor{\bsnm{Breunig}, \binits{M.M.}},
\bauthor{\bsnm{Kriegel}, \binits{H.-P.}},
\bauthor{\bsnm{Ng}, \binits{R.T.}},
\bauthor{\bsnm{Sander}, \binits{J.}}:
\batitle{Lof: Identifying density-based local outliers}.
\bjtitle{SIGMOD Rec.}
\bvolume{29}(\bissue{2}),
\bfpage{93}--\blpage{104}
(\byear{2000})
\doiurl{10.1145/335191.335388}
\end{barticle}
\endbibitem

\bibitem[\protect\citeauthoryear{Tang et~al.}{2002}]{COF}
\begin{bchapter}
\bauthor{\bsnm{Tang}, \binits{J.}},
\bauthor{\bsnm{Chen}, \binits{Z.}},
\bauthor{\bsnm{Fu}, \binits{A.W.-C.}},
\bauthor{\bsnm{Cheung}, \binits{D.W.-L.}}:
\bctitle{Enhancing effectiveness of outlier detections for low density patterns}.
In: \bbtitle{Pacific-Asia Conference on Knowledge Discovery and Data Mining}
(\byear{2002})
\end{bchapter}
\endbibitem

\bibitem[\protect\citeauthoryear{Bischl et~al.}{2025}]{OpenML2013}
\begin{barticle}
\bauthor{\bsnm{Bischl}, \binits{B.}},
\bauthor{\bsnm{Casalicchio}, \binits{G.}},
\bauthor{\bsnm{Das}, \binits{T.}},
\bauthor{\bsnm{Feurer}, \binits{M.}},
\bauthor{\bsnm{Fischer}, \binits{S.}},
\bauthor{\bsnm{Gijsbers}, \binits{P.}},
\bauthor{\bsnm{Mukherjee}, \binits{S.}},
\bauthor{\bsnm{Müller}, \binits{A.C.}},
\bauthor{\bsnm{Németh}, \binits{L.}},
\bauthor{\bsnm{Oala}, \binits{L.}},
\bauthor{\bsnm{Purucker}, \binits{L.}},
\bauthor{\bsnm{Ravi}, \binits{S.}},
\bauthor{\bsnm{{van Rijn}}, \binits{J.N.}},
\bauthor{\bsnm{Singh}, \binits{P.}},
\bauthor{\bsnm{Vanschoren}, \binits{J.}},
\bauthor{\bsnm{{van der Velde}}, \binits{J.}},
\bauthor{\bsnm{Wever}, \binits{M.}}:
\batitle{Openml: Insights from 10 years and more than a thousand papers}.
\bjtitle{Patterns}
\bvolume{6}(\bissue{7}),
\bfpage{101317}
(\byear{2025})
\doiurl{10.1016/j.patter.2025.101317}
\end{barticle}
\endbibitem

\bibitem[\protect\citeauthoryear{Lloyd}{1982}]{kmeans1}
\begin{barticle}
\bauthor{\bsnm{Lloyd}, \binits{S.}}:
\batitle{Least squares quantization in pcm}.
\bjtitle{IEEE transactions on information theory}
\bvolume{28}(\bissue{2}),
\bfpage{129}--\blpage{137}
(\byear{1982})
\end{barticle}
\endbibitem

\bibitem[\protect\citeauthoryear{Ankerst et~al.}{1999}]{ankerst1999optics}
\begin{barticle}
\bauthor{\bsnm{Ankerst}, \binits{M.}},
\bauthor{\bsnm{Breunig}, \binits{M.M.}},
\bauthor{\bsnm{Kriegel}, \binits{H.-P.}},
\bauthor{\bsnm{Sander}, \binits{J.}}:
\batitle{Optics: Ordering points to identify the clustering structure}.
\bjtitle{ACM Sigmod record}
\bvolume{28}(\bissue{2}),
\bfpage{49}--\blpage{60}
(\byear{1999})
\end{barticle}
\endbibitem

\bibitem[\protect\citeauthoryear{Frey and Dueck}{2007}]{affinityprop}
\begin{barticle}
\bauthor{\bsnm{Frey}, \binits{B.J.}},
\bauthor{\bsnm{Dueck}, \binits{D.}}:
\batitle{Clustering by passing messages between data points}.
\bjtitle{Science}
\bvolume{315}(\bissue{5814}),
\bfpage{972}--\blpage{976}
(\byear{2007})
\doiurl{10.1126/science.1136800}
{\href{https://arxiv.org/abs/https://www.science.org/doi/pdf/10.1126/science.1136800}{{https://www.science.org/doi/pdf/10.1126/science.1136800}}}
\end{barticle}
\endbibitem

\bibitem[\protect\citeauthoryear{Ester et~al.}{1996}]{dbscan}
\begin{bchapter}
\bauthor{\bsnm{Ester}, \binits{M.}},
\bauthor{\bsnm{Kriegel}, \binits{H.-P.}},
\bauthor{\bsnm{Sander}, \binits{J.}},
\bauthor{\bsnm{Xu}, \binits{X.}}:
\bctitle{A density-based algorithm for discovering clusters in large spatial databases with noise}.
In: \bbtitle{Proceedings of the Second International Conference on Knowledge Discovery and Data Mining}
(\byear{1996})
\end{bchapter}
\endbibitem

\bibitem[\protect\citeauthoryear{Sculley}{2010}]{minibkmeans}
\begin{bchapter}
\bauthor{\bsnm{Sculley}, \binits{D.}}:
\bctitle{Web-scale k-means clustering}.
In: \bbtitle{Proceedings of the 19th International Conference on World Wide Web}.
\bsertitle{WWW '10},
pp. \bfpage{1177}--\blpage{1178}.
\bpublisher{Association for Computing Machinery},
\blocation{New York, NY, USA}
(\byear{2010})
\end{bchapter}
\endbibitem

\bibitem[\protect\citeauthoryear{Benavoli et~al.}{2017}]{Ropetutorial}
\begin{barticle}
\bauthor{\bsnm{Benavoli}, \binits{A.}},
\bauthor{\bsnm{Corani}, \binits{G.}},
\bauthor{\bsnm{Dem{\v{s}}ar}, \binits{J.}},
\bauthor{\bsnm{Zaffalon}, \binits{M.}}:
\batitle{Time for a change: a tutorial for comparing multiple classifiers through bayesian analysis}.
\bjtitle{Journal of Machine Learning Research}
\bvolume{18}(\bissue{77}),
\bfpage{1}--\blpage{36}
(\byear{2017})
\end{barticle}
\endbibitem

\bibitem[\protect\citeauthoryear{Dem{\v{s}}ar}{2006}]{CD}
\begin{barticle}
\bauthor{\bsnm{Dem{\v{s}}ar}, \binits{J.}}:
\batitle{Statistical comparisons of classifiers over multiple data sets}.
\bjtitle{Journal of Machine Learning Research}
\bvolume{7}(\bissue{1}),
\bfpage{1}--\blpage{30}
(\byear{2006})
\end{barticle}
\endbibitem

\bibitem[\protect\citeauthoryear{Vinh et~al.}{2010}]{JMLR:metrics}
\begin{barticle}
\bauthor{\bsnm{Vinh}, \binits{N.X.}},
\bauthor{\bsnm{Epps}, \binits{J.}},
\bauthor{\bsnm{Bailey}, \binits{J.}}:
\batitle{Information theoretic measures for clusterings comparison: Variants, properties, normalization and correction for chance}.
\bjtitle{Journal of Machine Learning Research}
\bvolume{11}(\bissue{95}),
\bfpage{2837}--\blpage{2854}
(\byear{2010})
\end{barticle}
\endbibitem

\bibitem[\protect\citeauthoryear{Amos et~al.}{2023}]{icnn}
\begin{bchapter}
\bauthor{\bsnm{Amos}, \binits{B.}},
\bauthor{\bsnm{Cohen}, \binits{S.}},
\bauthor{\bsnm{Luise}, \binits{G.}},
\bauthor{\bsnm{Redko}, \binits{I.}}:
\bctitle{Meta optimal transport}.
In: \bbtitle{ICML 2023}
(\byear{2023})
\end{bchapter}
\endbibitem

\bibitem[\protect\citeauthoryear{Blei et~al.}{2003}]{lda}
\begin{barticle}
\bauthor{\bsnm{Blei}, \binits{D.M.}},
\bauthor{\bsnm{Ng}, \binits{A.Y.}},
\bauthor{\bsnm{Jordan}, \binits{M.I.}}:
\batitle{Latent dirichlet allocation}.
\bjtitle{J. Mach. Learn. Res.}
\bvolume{3}(\bissue{null}),
\bfpage{993}--\blpage{1022}
(\byear{2003})
\end{barticle}
\endbibitem

\end{thebibliography}
\newpage
\appendix
\section{Ablations of preprocessing functions}
To evaluate the role of preprocessing in our framework, we implemented and compared two alternative techniques: Latent Dirichlet Allocation (LDA)~\cite{lda} and Principal Component Analysis (PCA). While LDA occasionally produced favorable results depending on the initialization, its inherently stochastic nature led to inconsistent performance across runs, making it unsuitable for our meta-learning setting, which requires stable representations for comparison. PCA, on the other hand, resulted in numerically unstable outputs when paired with the Gromov-Wasserstein (GW) distance, occasionally leading to failed or ill-conditioned computations.

Although a deeper investigation into the interaction between PCA and GW distance may be worthwhile, it is beyond the scope of the current work. We appreciate that this could be an interesting direction for future studies and have chosen FastICA as the default preprocessing method due to its empirical robustness in our setting.

\section{Computational Complexity of Meta-Testing Phase}

We provide a detailed complexity analysis of the meta-testing phase of the \textsc{LOTUS} framework. Let:
\begin{itemize}
    \item $N$: number of datasets in the meta-dataset $\mathcal{D}_{\text{meta}}$
    \item $n$: number of samples in the test dataset
    \item $d$: number of features
    \item $r$: rank used in the low-rank approximation for Gromov-Wasserstein
\end{itemize}

The meta-testing pipeline involves the following key components:

\subsection*{1. FastICA Transformation}
FastICA is used to decorrelate and normalize the feature space. For a dataset with $n$ samples and $d$ features:
\[
\text{Cost per dataset: } \mathcal{O}(nd^2 + d^3)
\]
This includes PCA-based whitening ($\mathcal{O}(nd^2)$) and fixed-point ICA iterations ($\mathcal{O}(d^3)$). For $N$ meta-datasets, the total cost is:
\[
\mathcal{O}(N \cdot (nd^2 + d^3))
\]

\subsection*{2. Similarity Ranking via GWLR}
We compute the Gromov-Wasserstein distance between the test dataset and each of the $N$ meta-datasets using the low-rank GW formulation (Scetbon et al., 2021). The per-comparison cost is:  $\mathcal{O}(nr(r + d))$

Thus, the full similarity ranking step has complexity:
\[
\mathcal{O}(N \cdot nr(r + d))
\]

\subsection*{3. Overall Meta-Testing Complexity}
Combining the two stages, the total complexity is:
\[
\boxed{\mathcal{O}(N \cdot (nr(r + d) + nd^2 + d^3))} 
\]

This analysis shows that the meta-testing phase is linear in the number of reference datasets ($N$) and polynomial in the input dimensions and r and d are fixed and generally small.

\end{document}